%% file: main.tex
\documentclass{article}
\usepackage[left=3cm,right=3cm]{geometry}
\usepackage[utf8]{inputenc}
\usepackage[natbibapa]{apacite} 
\usepackage{hyperref} 
\usepackage{amsmath} 
\usepackage{amsfonts} 
\usepackage{authblk} 
\usepackage{setspace} 
\usepackage[colorinlistoftodos]{todonotes} 
\usepackage{caption}
\usepackage{subcaption}
\usepackage{xfrac}
\usepackage{booktabs}
\usepackage{todonotes}
\usepackage{doi} 

\usepackage{tikz} 
\usetikzlibrary{fit}
\usetikzlibrary{arrows.meta}
\tikzset{>={Stealth[scale=1]}}

\definecolor{bfblue}{RGB}{0, 91, 150}

\usepackage{orcidlink}

\hypersetup{colorlinks=true,allcolors=[rgb]{0,0,0.8}}

\newcommand{\applytoeach}[5]{\left\{ #1(#2_{#3}) \right\}_{#3=#4}^{#5}}

\DeclareMathOperator*{\argmin}{arg\,min}

\begin{document}
\title{Amortized Bayesian Mixture Models}

\author{Šimon Kucharský\orcidlink{0000-0003-4192-1140}}
\author{Paul-Christian Bürkner\orcidlink{0000-0001-5765-8995}}
\affil{Department of Computational Statistics, TU Dortmund University, Dortmund, Germany}
\date{}

\clearpage\maketitle

\begin{abstract}
    \noindent Finite mixtures are a broad class of models useful in scenarios where observed data is generated by multiple distinct processes but without explicit information about the responsible process for each data point. Estimating Bayesian mixture models is computationally challenging due to issues such as high-dimensional posterior inference and label switching. Furthermore, traditional methods such as MCMC are applicable only if the likelihoods for each mixture component are analytically tractable.
    
    Amortized Bayesian Inference (ABI) is a simulation-based framework for estimating Bayesian models using generative neural networks. This allows the fitting of models without explicit likelihoods, and provides fast inference. ABI is therefore an attractive framework for estimating mixture models. This paper introduces a novel extension of ABI tailored to mixture models. We factorize the posterior into a distribution of the parameters and a distribution of (categorical) mixture indicators, which allows us to use a combination of generative neural networks for parameter inference, and classification networks for mixture membership identification. The proposed framework accommodates both independent and dependent mixture models, enabling filtering and smoothing. We validate and demonstrate our approach through synthetic and real-world datasets. 
\end{abstract}


\newpage

\section{Introduction}

Fast and accurate estimation of statistical quantities is an ongoing problem in statistical research \citep{cranmer_frontier_2020,hermans_crisis_2022,papamakarios2016fast}. One major difficulty arises when the observed data is generated by multiple distinct processes, but the specific process responsible for each data point is unknown. Mixture models are commonly used to address this issue \citep{mclachlan_mixture_1988,fruhwirth-schnatter_finite_2006,zucchini_hidden_2016,visser_mixture_2022,scrucca_model-based_2023}.

Although mixture models have been useful across a wide range applications \citep[e.g.,][to name a few]{schaaf_hierarchical_2019,zavadskiy_functional_2024,kucharsky_hidden_2021,hadj-amar_bayesian_2023}, estimating mixture models can be challenging for several reasons. In practice, the most relevant obstacles  are: (1) obtaining full Bayesian inference for mixtures can be computationally demanding, (2) fitting Bayesian mixtures using standard methods (such as MCMC) requires the model likelihood to be analytically tractable, and (3) label switching issues, where the likelihood is invariant to permutations of mixture component labels, which makes the model identified only up to the permutation of the labels.

Given these challenges, there is a need for methods that can perform fast and accurate inference for mixture models while removing limitations of traditional methods. A promising candidate for resolving these issues is Amortized Bayesian Inference \citep[ABI, ][]{gershman_amortized_2014,radev_bayesflow_2022,ritchie_deep_2016,papamakarios2016fast}, which offers fast approximation of the posterior distributions even for models that are otherwise not analytically tractable. However, these methods have yet to be fully adapted to handle mixture models, in particular, when we desire estimating the parameters of the mixture model and the categorical latent mixture indicators as a joint probability distribution. This gap motivates the development of a new approach that uses modern deep learning techniques to enable scalable ABI for mixture models.

Our work provides the following contributions:

\begin{itemize}
    \item We develop a deep learning architecture that takes advantage of a factorization of mixture models where parameter posteriors can be estimated directly from data, and mixture membership classification is estimated based on data and the parameter estimates. This allows full Bayesian inference providing the joint posterior distribution of all quantities of interest.
    \item The implementation is amortized, meaning that the inference is considerably faster than traditional -- non-amortized -- methods, such as MCMC or other Approximate Bayesian Computation methods.
    \item The amortized mixture implementation extends amortized posterior estimation, which is typically limited to continuous quantities, with the amortized estimation of categorical latent variables with bounded cardinality.
    \item The proposed framework is able to handle independent mixture models as well as dependent mixtures. In the case of dependent mixtures, our method is able to perform both filtering: inferring the latent state at each time given observations up to that time, and smoothing: inferring latent states using the full sequence of observations.
    \item We validate our method on a battery of toy and real world examples. Where possible, we compare our results to that of \texttt{Stan} \citep{carpenter_stan_2017} as a gold standard for MCMC estimation.
    \item The method is implemented in a \texttt{Python} \citep{rossum_python_2010} library \texttt{BayesFlow} \citep{radev_bayesflow_2022} that provides user-friendly interface for ABI. 
\end{itemize}

The article is organized as follows. Section~\ref{sec:methods} describes the methods used and developed in this article. Section~\ref{sec:bmm} explains Bayesian mixture models and describes challenges when fitting them with traditional methods. Section~\ref{sec:sbi} and Section~\ref{sec:abi} provide a primer explaining existing methods for simulation-based amortized inference that our approach is built upon. Section~\ref{sec:nemm} describes methods developed in this article that implement amortized estimation of mixture models, and Section~\ref{sec:alternatives} discusses possible alternative approaches. Section~\ref{sec:case-studies} presents applications of our method in three case studies - two simple toy examples to showcase how the approach works for independent and dependent mixtures, and one empirical study that showcases inferences on real data. We summarize our work in Section~\ref{sec:conclusion}.

\section{Methods}
\label{sec:methods}

In this section, we introduce our proposed ABI framework for Bayesian mixture models. First, we will define Bayesian mixture models and highlight challenges that arise when fitting them with traditional, non-amortized methods. Second, we will explain the general idea behind simulation based inference and ABI. Then, we will explain how ABI can be extended to provide amortized inference for mixture models.

\subsection{Bayesian mixture models}
\label{sec:bmm}

We define Bayesian mixture models as a joint distribution $p(y, z, \theta)$: $y \in \mathcal{Y}$ represents all relevant observable variables, whereas $z \in \mathbb{N}$ and $\theta \in \mathcal{R}$ represent typically latent (unobservable) quantities; $z$ represents the latent mixture indicators, while $\theta$ represents parameters of the mixture model: the mixture weights as well as parameters required by all mixture components. We focus on situations where the model is assumed (or known) to consist of a mixture of $K$ populations (or processes), each with their own model $p(y_i \mid z_i, \theta)$ for the observable data $y_{1,\dots,N}$, each associated with a latent mixture membership indicator $z_{1,\dots,N} \in \{1, \dots, K\}$.

We assume that a generative model can be created by factorizing the joint distribution of the mixture model into a prior $p(z, \theta)$ and a likelihood $p(y \mid z, \theta)$: $p(y, z, \theta) = p(z, \theta)p(y \mid z, \theta)$. Further, we assume that it is possible to obtain random samples of $z^{(s)}$ and $\theta^{(s)}$ from the prior, and it is possible to obtain random samples of synthetic data $y^{(s)}$ from the likelihood, conditionally on $z^{(s)}$ and $\theta^{(s)}$. The generative model can be therefore implemented as a computer program which samples the triple $(z^{(s)}, \theta^{(s)}, y^{(s)})$ as follows:

\begin{equation}
    \begin{aligned}
        z^{(s)}, \theta^{(s)} & \sim p(z,\theta) \\ 
        y^{(s)} & \sim p(y \mid z^{(s)}, \theta^{(s)}).
    \end{aligned}
\end{equation}

Typically, most mixture models used in practice can be cast as a special case of this factorization:

\begin{equation}
    \begin{aligned}
        \theta^{(s)} & \sim p(\theta) \\
        z^{(s)} & \sim p(z \mid \theta^{(s)}) \\ 
        y^{(s)} & \sim p(y \mid z^{(s)}, \theta^{(s)}).
    \end{aligned}
\end{equation}

The aim of Bayesian analysis is to estimate the joint posterior of all unobserved variables, $p(\theta, z \mid y)$. This joint posterior is notoriously too complex to calculate analytically \citep{diebolt_estimation_1994}. Thus, approximation methods are typically needed to estimate the posterior distribution \citep{fruhwirth-schnatter_markov_2001,marin_bayesian_2005}.

The estimation of Bayesian mixture models is generally achieved using one of two common approaches, somewhat analogically to using ``classification'' vs. ``mixture'' likelihoods for fitting mixture models with maximum likelihood estimation \citep{mclachlan_classification_1982,ganesalingam_classification_1989}.

Conceptually, the first approach samples from the joint posterior of the model parameters $\theta$ and latent indicators $z$ directly:
\begin{equation}
\label{eq:direct_mcmc}
    \begin{aligned}
        p(\theta, z \mid y) \propto p(\theta) \prod_{i=1}^N p(z_i \mid \theta) p(y_i \mid z_i, \theta).
    \end{aligned}
\end{equation}
Such approach is typically implemented using MCMC with Gibbs sampling or its extensions \citep{diebolt_estimation_1994,marin_bayesian_2005,celeux_computational_2000}. The obvious computational obstacle is that given a set of $N$ data points each generated from one of $K$ processes, there is $K^N$ possible mixture membership permutations and it may be extremely difficult to sample from this distribution \citep{marin_bayesian_2005}. The MCMC sampler may venture into a set of ``trapping states'' from which it may take an enormous number of steps to escape from \citep{marin_bayesian_2005,diebolt_estimation_1994,celeux_computational_2000}. Although sampling from the full joint distribution is feasible for relatively small sample sizes and number of mixture components using MCMC samplers, such approach tends to scale badly with increasing sample size and with more mixture components. 

Probabilistic programming languages such as \texttt{Stan} \citep{carpenter_stan_2017} do not allow such implementation in the first place, because gradient-based sampling methods like Hamiltonian Monte Carlo require continuous variables for evaluating the gradients. As such, using discrete parameters directly in the MCMC is not allowed. Instead, models with discrete parameters must be handled with alternative approaches. The most common alternative is to factorize the joint posterior as $p(\theta, z \mid y) = p(\theta \mid y) p(z \mid \theta, y)$; first we sample from the posterior distribution of $\theta$ with $z$ marginalized out, 
\begin{equation}
\label{eq:marginalized_mcmc}
    \begin{aligned}
        p(\theta \mid y) & \propto p(\theta) \prod_{i=1}^N p(y_i \mid \theta),
    \end{aligned}
\end{equation}
where $p(y_i \mid \theta) = \sum_{k=1}^K p(z_i = k \mid \theta) p(y_i \mid z_i=k, \theta)$ is the likelihood of the observations with the mixture indicators being marginalized out. Subsequently, the distribution of $p(z \mid \theta, y)$ can be fully determined conditionally on the parameters $\theta$ and observations $y$:
\begin{equation}
    \begin{aligned}
        p(z_i \mid y_i, \theta) & = \frac{p(z_i \mid \theta) p(y_i \mid z_i, \theta)}{p(y_i \mid \theta)}.
    \end{aligned}
\end{equation}
Both factorizations described above are valid only if the latent indicators are independent and observations are conditionally independent given latent indicators (i.e., in the case of exchangeable data). In many cases, this assumption is not satisfied. For example, in hidden Markov models \citep[HMMs;][]{rabiner_tutorial_1989}, the observables are indeed independent conditionally on the latent indicators, but the latent indicators themselves form a Markov chain: the probability of the current state depends on the previous state(s). Other types of mixture models might exhibit other kinds of dependencies between observables or states -- depending on the exact nature of these dependencies, other forms of factorizations might be possible \citep[for examples, see][]{may_spatial_2024,same_clustering_2020,ambroise_clustering_1997,hadj-amar_bayesian_2023}. 

In any case, estimating mixture parameters and mixture memberships requires evaluating the likelihood density. As a result, the likelihood has to be analytically tractable under each mixture component for applicability of the above described density-based approaches.



\subsection{Simulation-Based Inference}
\label{sec:sbi}

Statistical inference requires specifying the likelihood function $p(y \mid \theta)$ that describes the link between parameters and data. However, scientific models are often formulated only as a \emph{simulation program} that may render the likelihood analytically intractable \citep{cranmer_frontier_2020}, for example, because the model involves differential equations without analytical solutions, or other complex generative procedures \citep[e.g.,][]{radev_bayesflow_2022,brehmer_simulation-based_2021,lueckmann_benchmarking_2021,boelts_flexible_2022}.

The Bayesian model is then available only as a probabilistic generative model of the triple of the prior $p(\theta)$ for model parameters $\theta$, a stochastic model $p(\nu \mid \theta)$ for nuisance variables (i.e., noise) $\nu$, and a \emph{simulation program} $g: (\theta, \nu) \rightarrow y$ that generates synthetic data $y$. The complete forward model can be defined as

\begin{equation}
    y = g(\theta, \nu) \text{ with }\nu \sim p(\nu \mid \theta), \theta \sim p(\theta).
\end{equation}

We can sample from this stochastic model repeatedly to obtain the pairs of data-generating parameters $\theta$ along with the observable data $y$. The likelihood is implicitly defined as an integral over all possible execution paths of the generative model (represented by the stochastic variables $\nu$),

\begin{equation}
\label{eq:implicit-likelihood}
    p(y \mid \theta) = \int p(y, \nu \mid \theta) d\nu,
\end{equation}
even though the analytic solution to that integral may be unknown and therefore an explicit analytic form of the likelihood $p(y \mid \theta)$ unavailable \citep{cranmer_frontier_2020}. 

Estimating statistical models with intractable likelihoods is commonly referred to as \emph{likelihood-free inference}, even though as presented in Equation~\ref{eq:implicit-likelihood}, the likelihood function exists, albeit it may be implicit. A more apt designation of inference without explicit analytic likelihoods is \emph{simulation-based inference} \citep[SBI,][]{cranmer_frontier_2020}, since these approaches rely on using (often extensive) Monte Carlo simulations from the generative model to perform inference on the model parameters.

A downside of SBI methods is computational complexity. When the likelihood is tractable, density-based methods (e.g., MCMC) are often preferred over SBI, because using the likelihood directly typically requires less computational resources to achieve comparable accuracy \citep{schmitt_consistency_2024,zeghal_neural_2022,brehmer_mining_2020}.

\subsection{Amortized Bayesian Inference}
\label{sec:abi}

Computational complexity remains a significant challenge for methods that approximate posterior distributions \citep{papamakarios2016fast}. When posterior inference must be performed repeatedly, such as when applying a model to multiple datasets or during model evaluation procedures like simulation-based calibration \citep{talts_validating_2018} and cross-validation \citep[see e.g.,][]{vehtari_practical_2017,burkner_approximate_2020}, the cost of standard SBI methods but also MCMC can quickly become prohibitive. In such cases, standard methods may be computationally infeasible and thus impractical.

Amortized Bayesian Inference \citep{gershman_amortized_2014,ritchie_deep_2016,radev_bayesflow_2022,papamakarios2016fast,gonccalves2020training} is a solution to both the problem of implicit likelihoods, and the problem of computational demands for inference. ABI divides model fitting into two distinct stages. In the first --- \emph{training} --- stage, neural networks learn the posterior based on simulated data from the generative model. In the second --- \emph{inference} --- stage, given any observed data $y^\text{obs}$, samples from the posterior distribution are simply obtained by generating samples from the trained inference network.  Most of the computational resources are expended during the training stage, allowing us to \emph{amortize} (pay upfront) the cost of inference, making subsequent fitting of the model during the inference stage substantially more effective.

Using neural networks to approximate posterior distributions is often referred to as neural posterior estimation (NPE). In this approach, the target posterior $p(\theta \mid y)$ is represented by a surrogate density $q_\phi(\theta \mid y)$, parametrized by a set of learnable network weights $\phi$ of an \emph{inference network} $f_\phi$. Such network is often implemented as a normalizing flow \citep{rezende2015variational,dinh2016density,kobyzev_normalizing_2020,papamakarios_normalizing_2021}. Other ML generative models, such as diffusion models \citep{sharrock_sequential_2024}, flow matching \citep{lipman2022flow,dax_flow_2023}, consistency models \citep{schmitt_consistency_2024}, and others, can be used for this purpose, and would be compatible with the methods developed in this article. Given that normalizing flows have demonstrated strong performance across various disciplines \citep[e.g.,][]{von_krause_mental_2022,schumacher_validation_2024,schumacher_neural_2023}, we focus on normalizing flows as one example of NPE.

Normalizing flows \citep{kobyzev_normalizing_2020,dinh2016density,rezende2015variational} are implemented using conditional invertible neural networks \citep[CINNs,][]{ardizzone_guided_2019}. The flow serves as a learnable transformation $f_\phi$ between the posterior (target) distribution and a base distribution, conditioned on the data. The transformation $f_\phi$ transforms the (potentially intractable) target distribution into this base distribution. The base distribution is typically chosen to be simple and tractable, for which both density evaluation and sampling are easy and efficient, e.g., the Gaussian. Hence the name ``normalizing flow'': $f_\phi$ \emph{normalizes} the potentially intractable target distribution to a multivariate Gaussian; if $\theta \sim q_\phi(\theta \mid y)$, then $\xi = f_\phi(\theta;\ y) \sim \mathrm{MvN}(0, \mathbb{I})$.

To sample from the target distribution, we start with the base distribution and reverse the transformation. We first obtain $S$ samples from the base distribution $(\xi^{(1)}, \xi^{(2)}, \dots, \xi^{(S)})$. Then, we pass these samples through the transformation inverse $f^{-1}_\phi$,
\begin{equation}
\label{eq:normalizing_flow_sampling}
    \theta^{(s)} = f_{\phi}^{-1}\big(\xi^{(s)}; y\big) \text{ for } s \in \{ 1, \dots, S\}.
\end{equation}

Alternatively to sampling, the target density for some point $\theta$ is available by evaluating the corresponding density of the base distribution at the point $\xi = f_{\phi}\big(\theta;\ y\big)$, adjusted for the transformation $f_{\phi}$ via the change-of-variables formula,
\begin{equation}
\label{eq:normalizing_flow_density}
    q_{\phi}\big(\theta \mid y\big) = p_\text{base}\Big(\xi = f_{\phi}\big(\theta;\ y\big)\Big) \left| \det \frac{\partial f_{\phi}\big(\theta; y\big)}{\partial \theta}\right|.
\end{equation}

Architectures of normalizing flows are specifically designed such that these operations (forward and inverse transform, evaluating the Jacobian adjustment) are computationally efficient but at the same time the transform $f_\phi$ sufficiently expressive \citep{durkan2019neural,kobyzev_normalizing_2020,dinh2016density,rezende2015variational}.

To ensure that the network accurately captures the target posterior, it is trained to minimize the Kullback-Leibler (KL) divergence between the approximated posterior $q_\phi(\theta \mid y)$ and the true posterior $p(\theta \mid y)$. The optimization objective of the inference network can therefore be written as,
\begin{equation}
\begin{aligned}
\label{eq:abi:loss}
    \widehat{\phi} & = \argmin_\phi \mathbb{E}_{(\theta,y)\sim p(\theta,y)} \bigg[\text{KL}\Big(p(\theta\mid y) \mid \mid q_\phi(\theta\mid y)\Big)\bigg] \\
    & \propto \argmin_\phi \mathbb{E}_{(\theta,y)\sim p(\theta,y)} \Big[-\log q_\phi(\theta \mid y)\Big] \\
    & \approx \argmin_\phi \frac{1}{M}\sum_{m=1}^M -\log q_\phi(\theta^{(m)} \mid y^{(m)}).
\end{aligned}    
\end{equation}
The true posterior density $p(\theta \mid y)$ can be dropped from the equation because it is constant with respect of the learnable network weights $\phi$. The expectation is approximated using a Monte-Carlo estimate over $M$ draws generated from the Bayesian model $(\theta^{(m)}, y^{(m)}) \sim p(\theta, y)$. The simulated data $y^{(m)}$ serve as conditioning input to the inference network, which outputs the corresponding approximate posterior densities $q_\phi(\theta^{(m)} \mid y^{(m)})$ according to Eq.~\ref{eq:normalizing_flow_density}. The network weights $\phi$ are optimized by minimizing the negative log-density of the simulated parameters conditionally on the simulated data (Eq.~\ref{eq:abi:loss}). Because the network is trained on \emph{simulated} parameter-observations pairs $(\theta^{(m)}, y^{(m)})$, ABI falls under the umbrella of SBI.

In addition to the \emph{inference network}, ABI approaches can be expanded with \emph{summary networks}. A summary network compresses raw data $y$ into summary statistics $h_\psi(y)$, i.e., a lower-dimensional representation of the data (embeddings). This simplifies the task of the \emph{inference network} that is therefore concerned with an inference conditioned on a smaller number of informative inputs rather than on a (potentially large) number of individual data points. The architecture of the summary network needs to reflect the structure of the data; for example, permutation invariant networks are suitable to summarize exchangeable data, recurrent neural networks are suitable for time-series data, and so forth. These networks can typically also take input of various size but their output is of fixed length. This allows the ABI amortize over different designs (e.g., varying sample sizes). The inference and summary network are trained concurrently using the modified loss,

\begin{equation}
\label{eq:loss_npe}
\begin{aligned}
    \widehat{\phi}, \widehat{\psi} & = \argmin_{\phi,\psi} \mathbb{E}_{(\theta,y)\sim p(\theta,y)} \Big[-\log q_\phi(\theta \mid h_\psi(y))\Big].
\end{aligned}    
\end{equation}

Given enough capacity of the summary network, the learned summary statistics are approximately sufficient with regards to the target of the posterior inference: Using the summary statistics does not alter the target posterior distribution, if swapped with the raw data: $p(\theta \mid y) = p(\theta \mid h(y))$ \citep{chen_neural_2021,radev_bayesflow_2022}. Optionally, we might use an additional penalty term that forces the summaries be distributed according to a specified distribution (e.g., a Gaussian). This regularizes the summary network and allows additional model checks, such as detecting \emph{simulation gaps} \citep[i.e., when the data used for inference differ significantly from the data seen during training;][]{kothe_detecting_2024}.

Once trained, the networks can be used for inference, that is, for posterior estimation given any observed data set $y^{\text{obs}}$. First, the data is passed through the trained summary network to obtain its embedding $h_{\widehat{\psi}}(y^{\text{obs}})$. Then, analogically to Eq~\ref{eq:normalizing_flow_sampling}, sample $S$ draws from the base distribution $(\xi^{(1)}, \xi^{(2)}, \dots, \xi^{(S)})$. The posterior draws $(\theta^{(1)}, \theta^{(2)}, \dots, \theta^{(S)})$ are obtained via the \emph{inverse pass} through the inference network,
\begin{equation}
\label{eq:parameter_sampling}
    \theta^{(s)} = f_{\widehat{\phi}}^{-1}\big(\xi^{(s)}; h_{\widehat{\psi}}(y^{\text{obs}})\big) \text{ for } s \in \{ 1, \dots, S\}.
\end{equation}
Alternatively, the approximate posterior density for any values of parameters $\theta$ is available analogically to Eq.~\ref{eq:normalizing_flow_density},
\begin{equation}
\label{eq:npe_density}
    p(\theta \mid y^{\text{obs}}) \approx q_{\widehat{\phi}}\big(\theta \mid h_{\widehat{\psi}}(y^{\text{obs}})\big) = p\bigg(\xi = f_{\widehat{\phi}}\Big(\theta; h_{\widehat{\psi}}\big(y^{\text{obs}}\big)\Big)\bigg) \left| \det \frac{\partial f_{\widehat{\phi}}\big(\theta; h_{\widehat{\psi}}\big(y^{\text{obs}}\big)\big)}{\partial \theta}\right|.
\end{equation}

It is this separation of \emph{training} and \emph{inference} that makes ABI such a promising method. Since sampling from the posterior distribution using ABI only requires generating samples from the base distribution and passing the data through the trained networks (Eq.~\ref{eq:parameter_sampling}), it is relatively fast compared to non-amortized methods during inference. Often times, models that take minutes, hours or even days to fit with non-amortized methods can take as little time as fractions of a second with amortized methods \citep{radev_bayesflow_2022,hermans_crisis_2022}. The efficiency of ABI during inference is particularly helpful in scenarios where data needs to be fit in real time, when many datasets need to be fit with the same model \citep[e.g.,][]{von_krause_mental_2022}, when the same data set needs to be fit many times \citep[e.g., cross-validation,][]{burkner_brms_2017} -- sometimes under different assumptions or processing steps \citep[e.g., for sensitivity analysis,][]{elsemuller_sensitivity-aware_2024}, or for validating the model on a large number of simulations \citep[e.g., simulation-based calibration,][]{talts_validating_2018}. In many cases, such uses of Bayesian models are borderline unfeasible with non-amortized methods \citep{hermans_crisis_2022}. ABI makes these procedures within reach in a matter of seconds. Since at no point during training or inference the model likelihood or priors need to be evaluated, this approach is also applicable in scenarios with analytically intractable likelihoods or priors.


\subsection{Neural estimation of Mixture Models}
\label{sec:nemm}

The goal is to estimate the joint posterior of the model parameters and the mixture indicators. In the following, we will work with the following factorization:
\begin{equation}
\label{eq:factorization}
    \begin{aligned}
    p(\theta, \{z_i\} \mid \{y_{ij}\}) = p(\theta \mid \{y_{ij}\}) p(\{z_i\} \mid \{y_{ij}\}, \theta).
    \end{aligned}
\end{equation}
Here we used the set notation to highlight that the number of observational units in $i = 1,\dots,N$ and the number of observations per unit $j = 1,\dots,P_i$ can vary across different datasets.

Using the factorization above, we separate the problem in two parts, one of which entails estimating the parameter posterior $p(\theta \mid \{y_{ij}\})$ that was already described in Section~\ref{sec:abi}. This posterior contains the distribution of all mixture parameters, that is, mixture weights and parameters of all mixture components. What remains is formulating a neural approximation of the second term $p(\{z_i\} \mid \{y_{ij}\}, \theta)$, which yields the distribution of mixture indicators. Since this is essentially a classification problem, common network architectures, such as the multilayer perceptron \citep[MLP, ][]{rosenblatt_perceptron_1958,baum_capabilities_1988,murtagh_multilayer_1991}, are adequate candidate architectures. However, because the number of observations per unit can vary, the input of the classification network must be brought into the same dimensions. For this reason, we introduce an (optional) summary network $h_\omega$ that creates summary embedding for each observational unit $\applytoeach{h_\omega}{y}{i}{1}{N}$ \citep[similar to the \emph{local} summary network used in amortized multilevel models,][]{habermann_amortized_2024}. The task of the local summary network is to extract relevant information about each observational unit. The choice of the local summary network depends on the structure of the data within the units. Independent observations within units may be summarized by permutation-invariant networks such as Deep Sets \citep{zaheer_deep_2017} or Set Transformers \citep{lee2019set}. Time-series or otherwise dependent measures within units might need to be summarized via networks that take dependencies into account, e.g., convolutional neural networks (CNNs; \citealp{lecun1998gradient}), recurrent neural networks (RNNs; \citealp{elman1990finding,hochreiter1997long}), Transformers \citep{vaswani2017attention}, and so on.

\begin{figure}[!t]
    \centering
    \begin{subfigure}{0.45\textwidth}
        \centering
        \begin{tikzpicture}
            \input{figures/dag}
        \end{tikzpicture}
        \caption{Exchangeable observational units (independent mixtures)}
    \end{subfigure}
    \begin{subfigure}{0.4\textwidth}
        \centering
        \begin{tikzpicture}
            \input{figures/dag}
            \draw[->,thick] (z1) -- (z2);
            \draw[->,thick] (z2) -- (z3);
        \end{tikzpicture}
        \caption{Hidden Markov model}
    \end{subfigure}%
    \begin{subfigure}{0.4\textwidth}
        \centering
        \begin{tikzpicture}
            \input{figures/dag}
            \node[draw,fit={(z1) (z2) (z3)}] {};
        \end{tikzpicture}
        \caption{Joint prior over mixture indicators}
    \end{subfigure}
    \caption{Examples of dependencies between observational units in mixture models. (a) Exchangeable observational units permit factorizing the distribution of mixture indicators as $p(z \mid y, \theta) = \prod_{i=1}^N p(z_i \mid y_i, \theta)$. For non-exchangeable observational units such as in (b) and (c), local decoding is used to factorize the joint distribution; \emph{filtering} as $p(z \mid y, \theta) = \prod_{i=1}^N p(z_i \mid y_1, \dots, y_i, \theta)$, or \emph{smoothing} as $p(z \mid y, \theta) = \prod_{i=1}^N p(z_i \mid y_1, \dots, y_N, \theta)$. Figure inspired by \citet{burkner_efficient_2021}.}
    \label{fig:dag}
\end{figure}
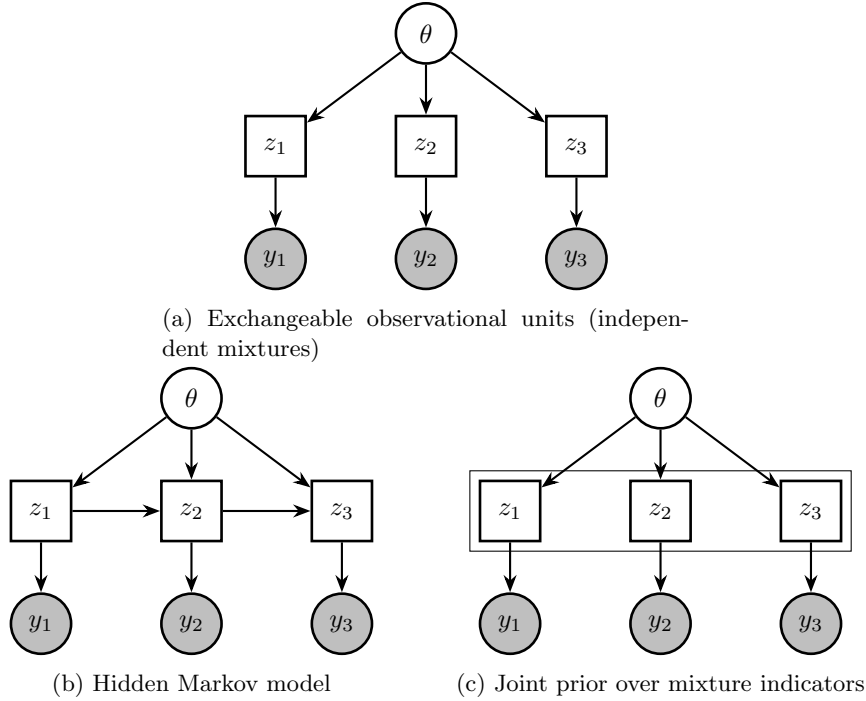

The exact architecture of the classification network depends on structure of the data, which dictates how to factorize the joint distribution $p(z \mid y, \theta)$. Some examples of data structures are shown in Figure~\ref{fig:dag}. In the case of exchangeable observational units, the joint distribution can be factorized as $p(z \mid y, \theta) = \prod_{i=1}^N p(z_i \mid y_i, \theta)$. This means that the probability distribution of each mixture indicator $z_i$ can be approximated independently --- conditioned on the parameters $\theta$ and observation $y_i$ --- with a \emph{classification} network as follows,
\begin{equation}
    p(z_i \mid y_{i}, \theta) \approx q_\alpha(z_i \mid y_{i}, \theta) =  \text{softmax} \Big(f_\alpha\big(h_\omega(y_{i}), \theta\big)\Big) \text{ for } i \in \{1, \dots, N\},
\end{equation}
here $f_\alpha$ stands for a MLP with a series of hidden layers connected by learnable weights $\alpha$ followed by non-linear activations, and an output layer with $K$ nodes. The final softmax activation converts the output into a set of mutually exclusive probabilities for each class, essentially implementing soft classification \citep{wahba_soft_2002,liu_hard_2011}. Soft classification refers to the calculation of class probabilities rather than producing classification labels directly, which is typical for hard categorical classification.

When the observational units are not exchangeable (e.g., ordered time series, spatial data), the joint distribution $p(z \mid y, \theta)$ may not be factorized easily due to dependencies across observational units. In such cases, directly modeling the joint probability distribution of all mixture indicators could require complex and computationally expensive architectures \citep{mark_bayesian_2018}. In this article, we address this problem using \emph{local decoding}: we \emph{assume} that the distribution can be factorized by conditioning on other observational units. Rather than modeling the full joint probability distribution of all mixture indicators \citep{viterbi_error_1967,lember_estimation_2019}, we approximate each indicator’s probability distribution one at a time, significantly simplifying the computational process \citep{sarkka_bayesian_2023}.

One method of local decoding is \emph{filtering} where the probability distribution of mixture indicator $z_t$ is based on the sequence of observations up until the $t^{th}$ data point, denoted as $\{y_i\}_{i=1}^{t}$,
\begin{equation}
    p(z_t \mid \{y_{i}\}_{i=1}^t, \theta) \approx q_\alpha(z_t \mid \{y_{i}\}_{i=1}^t, \theta) = \text{softmax} \Big(f_\alpha\big(\applytoeach{h_\omega}{y}{i}{1}{t}, \theta\big)\Big) \text{ for } t \in \{1, \dots, N\},
\end{equation}
where $f_\alpha$ is a \emph{forward} network that takes the set of observations (or their embeddings) and outputs activation pertaining to the last element in the set. A forward network can be any network that captures dependencies between observations, such as recurrent neural networks (RNNs), gated recurrent units (GRUs), transformers, an so forth. 

Both in the context of \emph{classification} and in the context of \emph{filtering}, the networks can be trained with realizations from the Bayesian generative model, where the tuple $(y, \theta)$ serves as an input, and $z$ is the predicted target. Traditional classification loss functions, such as categorical cross-entropy, are well-suited for this inference problem. Specifically, the following loss function can be used,
\begin{equation}
\label{eq:loss_foward}
    \widehat{\alpha}, \widehat{\omega} = \argmin_{\alpha,\omega} \mathbb{E}_{(\theta,y,z)\sim p(\theta,y,z)} \Big[ -\log q_\alpha\big(\{z_i\} \mid \{h_\omega(y_i)\}, \theta\big) \Big].
\end{equation}
An extension of \emph{filtering} is \emph{smoothing} \citep[qv,][]{sarkka_bayesian_2023,mark_bayesian_2018}, where the probability of a mixture indicator $z_t$ is based on all available observations instead of only the observations that precede the current data point. To compute the smoothing estimate, we combine the output of the forward network with an output of a \emph{backward} network $f_\beta$. The backward network is applied to the \emph{remaining} observations after $t$, and processes them in the reversed order denoted as $\{y_i\}_{i=N}^{t+1}$,
\begin{equation}
\label{eq:smoothing}
\begin{aligned}
    p(z_t \mid \{y_{i}\}_{i=1}^N, \theta) \approx q_{\alpha,\beta}(z_t \mid \{y_{i}\}_{i=1}^N, \theta) = \text{softmax} \Big(&f_\alpha\big(\applytoeach{h_\omega}{y}{i}{1}{t}, \theta\big) + \\&f_\beta\big(\applytoeach{h_\omega}{y}{i}{N}{t+1}, \theta\big)\Big)  \text{ for } t \in \{1, \dots, N\}.
\end{aligned}
\end{equation}
Although the forward and backward networks are, in principle, separate, in many cases it is justified for them to share the same weights such that $\alpha = \beta$. This is because both networks process the same type of information and have the same inferential target (the probability distribution of $z$). In some cases, the predictions from either the forward network or the backward network might systematically outperform the combined outputs of both networks. In that case, training the networks using the smoothed predictions in Eq.~\ref{eq:smoothing} could lead to a solution that effectively mimics the one network which performs better while ignoring the performance of the other network. To avoid such a scenario, instead of using the smoothed classification probabilities from Eq.~\ref{eq:smoothing}, both networks are trained with their own loss, each based on the classification probabilities in the forward and backward direction, respectively,
\begin{equation}
\begin{aligned}
\label{eq:loss_foward_backward}
    \widehat{\alpha}, \widehat{\beta}, \widehat{\omega} = \argmin_{\alpha,\beta,\omega} \mathbb{E}_{(\theta,y,z)\sim p(\theta,y,z)} \Big[ -&\log q_\alpha(\{z_i\} \mid \{h_\omega(y_i)\}, \theta) \\-&\log q_\beta(\{z_i\} \mid \{h_\omega(y_i)\}, \theta) \Big],
\end{aligned}
\end{equation}
to ensure that both networks are optimized for the classification task on their own. Smoothing in Eq.~\ref{eq:smoothing} is not part of the training loss. Instead, once the forward and backward networks have been trained separately (Eq.~\ref{eq:loss_foward_backward}), their outputs can be combined according to Eq.~\ref{eq:smoothing}  to yield smoothed classification probabilities. Thus, Eq.~\ref{eq:smoothing}  serves purely as a post-training prediction rule rather than a learning rule.

\begin{figure}[!t]
    \centering
    \begin{tikzpicture}
        \node[draw, thick, rectangle, minimum width=1.5cm, minimum height=1cm, fill=lightgray] (theta) at (2,2) {$\theta\sim p(\theta)$};
        \node[draw, thick, rectangle, minimum width=1.5cm, minimum height=1cm, fill=lightgray] (z) at (4,0) {$z_i \sim p(z \mid \theta)$};
        \node[draw, thick, rectangle, minimum width=1.5cm, minimum height=1cm, fill=lightgray] (y) at (0,0) {$y_i \sim p(y \mid z, \theta)$};

        \node[draw, thick, rectangle, dashed, minimum width=2cm, minimum height=1cm, align=center] (local) at (0,-2) {Local summary\\network};
    
        \node[draw, thick, rectangle, dashed, minimum width=2cm, minimum height=1cm, align=center] (global) at (-5,-2) {Global summary\\network};
        \node[draw, thick, rectangle, dashed, minimum width=2cm, minimum height=1cm, align=center] (invertible) at (-5,-4) {Posterior\\network};
        \node[draw, thick, rectangle, fill=bfblue, opacity=0.5, text opacity=1, align=center] (loss_posterior) at (-1,-4) {Loss:\\$-\log q_\phi\big(\theta \mid h_\psi(\{h_\omega(y)\})\big)$};
        
        \node[draw, thick, rectangle, dashed, minimum width=2cm, minimum height=1cm, align=center] (classification) at (4,-2) {Classification\\network};
        \node[draw, thick, rectangle, fill=bfblue, opacity=0.5, text opacity=1, align=center](loss_classification) at (4,-4) {Loss:\\$-\log q_\alpha\big(\{z_i\} \mid \{h_\omega(y_i)\}, \theta\big)$};

        \node[draw,fit={(y) (z) (local) (classification)},rounded corners=.2cm,inner sep=0.6cm,label={[anchor=north west]north west:$i \in \{1,\dots,N\}$}]    {};

        \draw[->] (theta) -- (y);
        \draw[->] (theta) -- (z);
        \draw[->] (z) to (y);
        
        \draw[->] (y) -- (local);
        \draw[->] (local) -- node[above, fill=white, fill opacity=0.9, text opacity=1, yshift=1pt, xshift=-7.5pt] {$\{h_\omega(y_i)\}$} (global);
        \draw[->] (global) -- node[right] {$h_\psi(\{h_\omega(y_i)\})$} (invertible);
        \draw[->] (theta) -- (-6.5, 2) -- (-6.5, -4) -- (invertible);
        \draw[->] (invertible) -- (loss_posterior);

        \draw[->] (theta) -- (5.5, 2) -- (5.5, -2) -- (classification);
        \draw[->] (local) -- node[above] {$h_\omega(y_i)$} (classification);
        \draw[->] (z) -- (classification);
        \draw[->] (classification) -- (loss_classification);
    \end{tikzpicture}
    \caption{Schematic representation of training amortized mixture models. The boxes highlighted in gray represent the inputs (i.e., the training set) sampled from the Bayesian generative model. The observations $y_i$ are individually passed through the local summary network. For parameter posterior training, the complete set of local summaries is further passed through the global summary network. The global summary is passed together with the true parameters $\theta$ to the posterior network to obtain the loss from Eq.~\ref{eq:loss_npe}. For classification training, the local summaries are concatenated with the true parameters $\theta$, and together passed with the true mixture indicators $z_i$ to the classification network, to obtain the loss from Eq.~\ref{eq:loss_foward} (or in case of separate forward and backward networks, Eq.~\ref{eq:loss_foward_backward}). Combining the two losses results in the joint loss in Eq.~\ref{eq:loss_combined}. The objective of training is to minimize the total loss by optimizing network weights $\phi, \psi, \omega, \alpha$. To simplify notation, we omitted indices $\theta^{(m)}$, $y_i^{(m)}$, $z_i^{(m)}$ indicating the training sample $m \in \{1,\dots,M\}$. }
    \label{fig:mixtures-schema-training}
\end{figure}
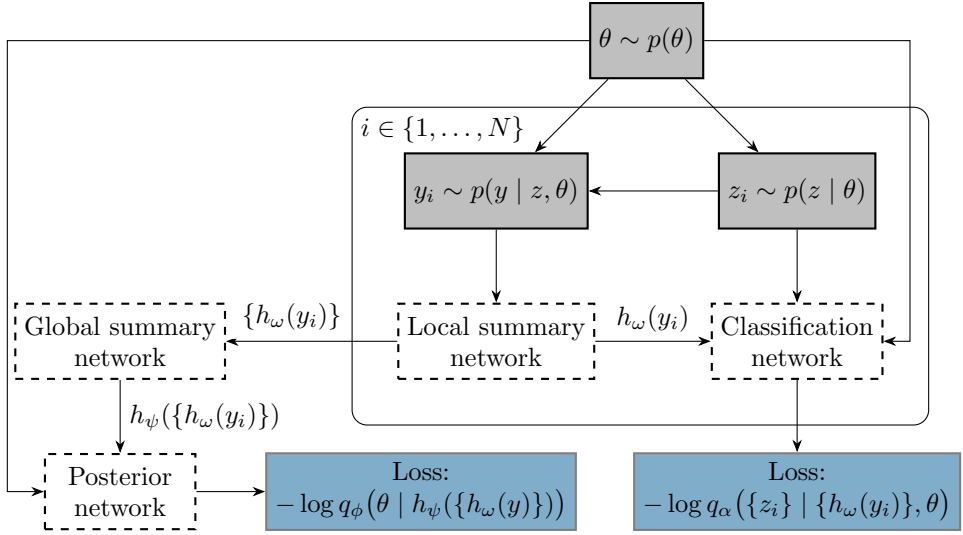

The losses in Eq.~\ref{eq:loss_npe} and Eq.~\ref{eq:loss_foward} or Eq.~\ref{eq:loss_foward_backward} can be combined, and all networks trained concurrently on the same set of training examples generated from the Bayesian generative model. Concurrent training also offers an opportunity for weight sharing between the NPE and classification tasks. Figure~\ref{fig:mixtures-schema-training} shows a schematic representation of joint training of the networks used for posterior and mixture inference.
For example, the local summary network $h_\omega$ can also be used during NPE to compress each observational unit before the global summary network $h_\psi$ is applied to compress the whole data set for posterior inference, leading to the following training objective involving the local ($h_\omega$) and global ($h_\psi$) summary networks, the posterior network ($q_\phi$), and the forward ($q_\alpha$) and backward ($q_\beta$) networks:
\begin{equation}
\label{eq:loss_combined}
\begin{aligned}
    \widehat{\phi}, \widehat{\psi}, \widehat{\alpha}, \widehat{\omega} = \argmin_{\phi,\psi,\alpha,\omega} \mathbb{E}_{(\theta,y,z)\sim p(\theta,y,z)} \Big[-&\log q_\phi\big(\theta \mid h_\psi(\{h_\omega(y_i)\}\big) \\
    -&\log q_\alpha(\{z_i\} \mid \{h_\omega(y_i)\}, \theta) \\
    -&\log q_\beta(\{z_i\} \mid \{h_\omega(y_i)\}, \theta)\Big].
\end{aligned}    
\end{equation}

\begin{figure}[!t]
    \centering
    \begin{tikzpicture}
        \node[draw, thick, rectangle, minimum width=1.5cm, minimum height=1cm, fill=lightgray] (y) at (0,0) {$y_i^{\text{obs}}$};
        
        \node[draw, thick, rectangle, dashed, minimum width=2cm, minimum height=1cm, align=center] (local) at (0,-2) {Local summary\\network};
    
        \node[draw, thick, rectangle, dashed, minimum width=2cm, minimum height=1cm, align=center] (global) at (-5,-2) {Global summary\\network};
        \node[draw, thick, rectangle, dashed, minimum width=2cm, minimum height=1cm, align=center] (invertible) at (-5,-4) {Posterior\\network};
        \node[draw, thick, rectangle, minimum width=1.5cm, minimum height=1cm, fill=bfblue, opacity=0.5, text opacity=1] (theta) at (5,-4) {$\theta^{(s)}$};

        \node[draw, thick, rectangle, dashed, minimum width=2cm, minimum height=1cm, align=center] (classification) at (5,-2) {Classification\\network};
        \node[draw, thick, rectangle, minimum width=1.5cm, minimum height=1cm, fill=bfblue, opacity=0.5, text opacity=1] (z) at (5,0) {$z_i^{(s)} $};
    
        \node[draw,fit={(y) (z) (local) (classification)},rounded corners=.2cm,inner sep=0.6cm,label={[anchor=north west]north west:$i \in \{1,\dots,N\}$}]    {};
        \node[draw,fit={(theta) (classification) (z)},rounded corners=.2cm,inner sep=0.8cm,label={[anchor=south west]south west:$s \in \{1,\dots,S\}$}]    {};

        \draw[->] (invertible) to node[below]{$q_\phi(\theta \mid y^{\text{obs}})$} (theta);

        \draw[->] (y) -- (local);
        \draw[->] (local) -- node[above,fill=white, fill opacity=0.9, text opacity=1, yshift=1pt] {$\{h_\omega(y_i^{\text{obs}})\}$} (global);
        \draw[->] (global) -- node[right] {$h_\psi(\{h_\omega(y_i^{\text{obs}})\})$} (invertible);

        \draw[->] (local) -- node[above,fill=white, fill opacity=0.9, text opacity=1, yshift=1pt, xshift=-9pt] {$h_\omega(y_i^{\text{obs}})$} (classification);

        \draw[->] (theta) -- (classification);
        \draw[->] (classification) to node[right, fill=white, fill opacity=0.9, text opacity=1, xshift=1pt]{$q_\alpha\big(z_i \mid y_i, \theta^{(s)}\big)$} (z);
    \end{tikzpicture}
    \caption{Schematic representation of the use of amortized mixture models for inference. The observations $y_i$ are individually passed through the local summary network. For parameter posterior inference, the complete set of local summaries is further passed through the global summary network. The global summary is passed to the posterior network to generate samples $\theta^{(s)}$ from the approximate posterior distribution. For classification, the local summaries are concatenated with the parameter samples $\theta^{(s)}$ and passed through the classification network to obtain the approximate mixture membership probabilities. If desired, the mixture indicators $z_i$ can be sampled from this approximate distribution.}
    \label{fig:mixtures-schema-inference}
\end{figure}
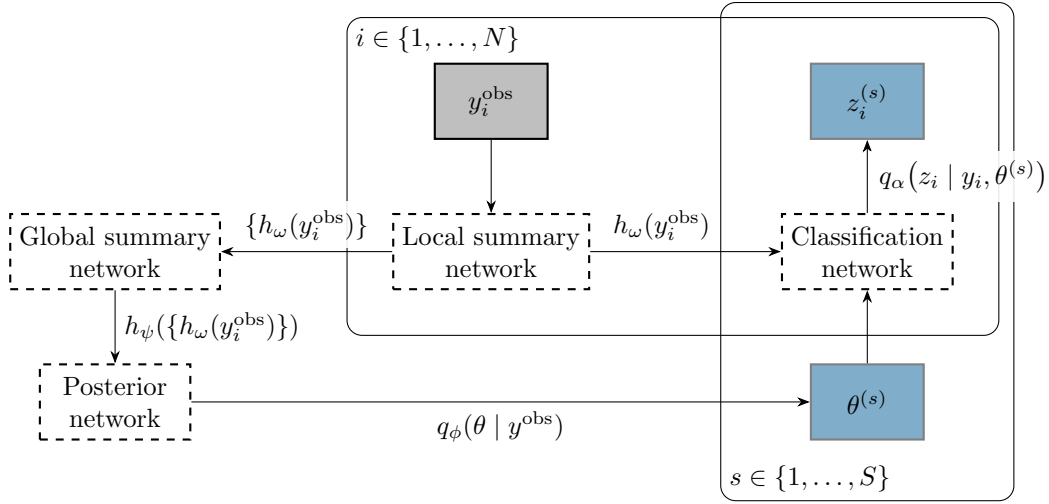

Once the networks are trained, they can be used to make fast inferences from any data. Figure~\ref{fig:mixtures-schema-inference} shows a schematic representation of using the networks for inference. The parameter posteriors are sampled as explained in Eq.~\ref{eq:parameter_sampling}. For the mixture classification, each sample from the posterior, $\theta^{(s)}$, is subsequently used in the mixture membership classification, leading to variations in the classification network output that is a result of the variability in the parameter values. This way, the uncertainty in parameter values is propagated to express the resulting uncertainty in mixture classification.

\subsection{Alternative factorizations}
\label{sec:alternatives}

The factorization used in Eq.~\ref{eq:factorization} reflects the approach described in Eq.~\ref{eq:marginalized_mcmc} where parameter posteriors are estimated first with the latent mixture indicators marginalized out (and subsequently recomputed). However, as explained in Section~\ref{sec:bmm}, this is not the only option to estimate mixture models. Instead, it would be possible to use an alternative factorization,
\begin{equation}
\label{eq:alternative_factorization}
p(\theta, \{z_i\} \mid \{y_i\}) = p(\{z_i\} \mid \{y_i\}) \, p(\theta \mid \{z_i\}, \{y_i\}),
\end{equation}
\noindent which would imply a different network architecture. Specifically, a classification network would need to predict mixture membership based on the (1) local and (2) global summary networks,
\begin{equation}
    p_\alpha\big(z_i \mid \{y_j\}_{j=1}^N\big) \approx \text{softmax}\bigg(f_\alpha\Big(h_\omega(y_i), h_\psi\big(\{h_\omega(y_j)_{j=1}^N\}\big)\Big)\bigg) \text{ for } i \in \{1, \dots, N\},
\end{equation}
which requires conditioning the classification network on the entire data set, as opposed to conditioning it on the parameters.

The mixture probability distribution $p(\{z_i\} \mid \{y_i\})$ by definition marginalizes out the parameter values, i.e., provides only the expectation of the probability distribution with respect to the parameter values, $p(\{z_i\} \mid \{y_i\}) = \mathbb{E}_{\theta \mid y}\big(p(\{z_i\} \mid \{y_i\}, \theta)\big)$. In principle, estimating the distribution of $\{z_i\}$ on its own would also be suitable for applications when parameter posteriors are not of interest, only mixture membership classification is. A notable example are Bayesian non-parametric mixtures with infinite number of components, where clustering of the data is the main focus of inference \citep[see][]{miller2018mixture}. Other applications of mixture models typically involve more detailed analysis of the joint probability distribution of $\theta$ and $\{z_i\}$. At minimum, parameter posteriors are used to check model fit or misspecification.

To estimate the joint distribution using the alternative factorization, the parameter posterior network would have to be informed by both the data and the latent indicators. One way how to achieve this is to introduce an additional global summary network $h_\xi$ that would learn embeddings of the concatenation of the local summary embeddings and the latent indicators,
\begin{equation}
    p(\theta \mid \{z_i\}, \{y_i\}) \approx q_\phi\Big(\theta \mid h_\xi\big(\{h_\omega(y_i), z_i\}\big)\Big).
\end{equation}

During inference, the mixture indicators would have to be sampled from a categorical distribution according to $\{z_i\} \sim p_\alpha(\{z_i\} \mid \{y_i\})$ for every sample $s$ from the parameter posterior. Note that if posterior samples of the mixture indicators are required, then draws from the categorical distribution are necessary even for the factorization used in this article. However, in most applications the quantities of interest are the mixture membership \emph{probabilities} rather than specific sampled realizations. Since these probabilities are provided directly by the classification network, additional sampling from the categorical distribution is not needed. Another disadvantage of this approach is the need to use two global summary networks -- one that is used for classification, and one that is used for posterior inference. For these reasons, we decided to use only the factorization described in our case studies in Section~\ref{sec:nemm}.

\subsection{Related work}

The present article puts a lot of emphasis on estimating the joint distribution of categorical and continuous variables -- in the present case, the latent mixture membership indicators, and the parameter of the mixture model. Modeling of continuous and discrete quantities was also proposed by \citet{boelts_flexible_2022}, used the same principle (factorizing the joint distribution) to model continuous and discrete data via neural networks (neural likelihood estimation). In the context of posterior inference, \citet{schroder2023simultaneous} proposed simultaneous inference over discrete model components and their parameters, essentially allowing model-averaged inference and prediction. The conceptual difference of this work is that whereas predicting model posterior requires a single classification conditioned on the whole dataset, inference for mixture models requires multivariate classification conditioned on individual observational units. The split between inferring \emph{local} and \emph{global} variables also makes the mixtures application a case related to multilevel and graphical models \citep{habermann_amortized_2024,arruda2025compositional,heinrich_hierarchical_2024}; in particular, the currently used factorization is conceptually similar to two-level multilevel models described by \citet{habermann_amortized_2024}.

\section{Case studies}
\label{sec:case-studies}

We evaluate our proposed approach for amortized Bayesian mixture models on three case studies. The first two case studies present idealized synthetic examples that demonstrate the use of our approach in the case of independent and dependent mixture models, respectively. The last example shows an application on a real world data set. All applications of ABI are implemented using the \texttt{BayesFlow} software package \citep{radev_bayesflow_2022} in \texttt{Python} \citep{rossum_python_2010}. Code associated with this article is accessible at \href{https://osf.io/7wvyk/}{osf.io/7wvyk/}.

\subsection{Evaluation}

An implementation of any model needs to be validated in order to ascertain whether conclusions made using the model are not misleading \citep{hermans_crisis_2022}. For each application, we apply a range of checks to validate the models we showcase in this article. We follow a principled Bayesian workflow \citep{schad_toward_2019,gelman_bayesian_2020,gabry_visualization_2019}.

Simulation-based calibration \citep[SBC,][]{talts_validating_2018,modrak_simulation-based_2023} is used to validate whether the parameter posterior approximation is well calibrated with respect to the \emph{true} posterior. In SBC, parameters are repeatedly sampled from the prior, synthetic data are generated from these parameters, and the model is refitted to each dataset (generate posterior draws). For each parameter, we then compute the rank of the true (data-generating) parameter value among its posterior draws. When the inferred posterior distribution is well calibrated with respect to the true data-generating process, these ranks follow a uniform distribution. To visualize this, we plot the fractional ranks: the normalized ranks divided by the number of posterior samples. Specifically, we show the difference between the expected ECDF of a uniform distribution and the observed ECDF of the fractional rank statistic. Ideally, the ECDF difference fluctuates close to zero. A failed SBC check indicates issues with computational validity of the parameter posterior approximation, which could be due to not sufficient training, or a lack of network expressiveness. See \citet{modrak_simulation-based_2023,talts_validating_2018} for more details about SBC. Additionally, parameter recovery, posterior \emph{z}-score, and posterior contraction are used to judge whether the data in combination with the statistical model lead to meaningful inferences. Poor results can indicate poor posterior calibration, but can also mean problems with parameter identifiability, or a lack of information provided by the data.

To detect model misspecification, we take several approaches. First, we adopt a method from the neural network literature that tests whether the observed data deviates from the typical data encountered during training \citep[i.e., \emph{simulation gaps};][]{kothe_detecting_2024}. Specifically, we use the Maximum Mean Discrepancy (MMD) as a discrepancy measure. During training, the global summary network is regularized with an MMD loss to ensure that the learned summary statistics approximately follow a Gaussian distribution. During inference, we compute the MMD between the empirical data and data simulated from the generative model, and compare it against a reference (null) distribution derived from the simulation model. This enables a formal test of whether the observed data are consistent with the data-generating process assumed during training \citep{kothe_detecting_2024}. From the Bayesian statistics perspective, we also perform \emph{posterior predictive checks} \cite[e.g.,][]{gelman_posterior_1996} to assess whether replicated data drawn from the posterior predictive distribution resemble the observed data. Since the first two case studies are demonstrations using synthetic data, the statistical models are by definition well specified. Therefore, we conduct model misspecification checks only in the final case study.

As a last step,  we investigate the accuracy of the neural inference by comparing it to results from state-of-the-art MCMC. To this end, models presented in our case studies were evaluated against the results using the probabilistic programming language \texttt{Stan} \citep{carpenter_stan_2017}. We compare posterior samples from \texttt{BayesFlow} and \texttt{Stan} to contrast our implementation of ABI to MCMC. Additionally, we also use the classifier two-sample test \citep[C2ST;][]{lopez2016revisiting} as a quantitative measure of distribution similarity. In our implementation of C2ST, ABI and MCMC posterior samples are labeled as separate classes and used to train a neural network classifier (MLP). The classifier's predictive performance is estimated via 5-fold cross-validation, and using classification accuracy as the evaluation metric. The resulting mean accuracy constitutes the C2ST score; values close to 0.5 indicate that the two sample sets are indistinguishable, whereas higher values suggest greater divergence between the underlying distributions.

\subsection{Parameter constraints}

Some parameters in the current models are naturally constraint -- for example, mixture proportions are bounded between zero and one and must sum to one. Across all experiments and applications, we ensured that all of the model parameters estimated by the amortized posterior estimator are trained on the unconstrained real space. Because the networks only see the unconstrained parameters during training, the parameter posterior network is subsequently making inferences on the unconstrained parameter space. Similarly, the classification networks are also conditioned on parameters in the unconstrained space. Where appropriate, we transform the parameters from the unconstrained space into the constrained space for easier interpretation of the results.
Here we explain the parameter transformations used across the case studies. 

\subsubsection*{Unit simplex}

Unit simplex parameters $\pi$ are constrained such that $0 \leq \pi_k \leq 1$ for $k\in\{1,\dots,K\}$ and $\sum_k^K \pi_k = 1$. Parameters constrained on the unit simplex in mixture models typically come in form of mixture proportions. Further, in hidden Markov models, each row of the latent state transition matrix is a unit simplex.
To explain our notation,
\begin{equation}
    \pi \sim \text{Dirichlet}(2, 2, 2)
\end{equation}
indicates that the elements of the vector $\pi$ are generated from a Dirichlet distribution with $K=3$ parameters. By construction, only $K-1$ parameters are needed to reconstruct the entire vector, since all values need to sum to one.

During training, the parameters are transformed into an unconstrained space $\theta \in \mathbb{R}^{K-1}$, defined for $1\leq k<K$,
\begin{equation}
    \theta_k = \text{logit}^{-1} \Bigg( \pi_k  + \log\Big(\frac{1}{K-k}\Big)\Bigg).
\end{equation}
During inference, the posterior parameter network outputs the posterior distribution on the unconstrained space. To convert the parameters back to the constrained space, the following transformation is applied,
\begin{equation}
    \pi_k = \begin{cases}
        \big(1 - \sum_j^{k-1} \theta_j\big) \theta_k & \text{if } 1 \leq k < K\\ 
        1 - \sum_{j=1}^{K-1} \pi_j & \text{if } k=K.
    \end{cases}
\end{equation}

\subsubsection*{Ordered vector}

A mixture model with the same distribution under each mixture component is identifiable up to the permutation of the mixture labels. To prevent label switching \citep[see][]{jasra_markov_2005}, one simple solution is to impose order constraints on parameters. While this naive approach can suffer from several drawbacks if used in isolation to solve the label switching problem \citep[e.g.,][]{celeux_computational_2000,marin_bayesian_2005}, we use order constraints in combination with (1) weakly informed priors (all case studies) and (2) differences between mixture components likelihoods (case study 3) to ensure model identifiability.
To explain the notation,
\begin{equation}
    (\mu_1, \mu_2, \mu_3) \sim \text{Normal}\Big((-2, 0, 2), \mathbb{I} \Big)_{\mu_1 < \mu_2 < \mu_3}
\end{equation}
indicates an order constraint of the mean parameters in a $K=3$ component mixture such that $\mu_1 < \mu_2 < \mu_3$. During sampling from the generative model, the order constraints are achieved by rejection sampling; first, we draw from the normal distribution, and if the draw does not satisfy the constraint the draw is repeated. In principle, rejection sampling is not an efficient way to draw from the constrained distribution. However, in our cases, the prior components are already relatively well separated, leading to low rejection rates. As a result, rejection sampling did not significantly slow down sampling from the Bayesian generative model.

During training, parameters are transformed to an unconstrained space, where
\begin{equation}
\begin{aligned}
    \theta_k = 
    \begin{cases}
        \mu_1 & \text{if } k=1 \\
        \log(\mu_{k} -\mu_{k-1}) & \text{if } 1 \leq k < K.
    \end{cases}
\end{aligned}
\end{equation}

During inference, the parameter posterior network returns the posterior distribution on the unconstrained space. To convert the parameters back into the constrained space, the following transformation is applied,
\begin{equation}
\begin{aligned}
    \mu_k = 
    \begin{cases}
        \theta_1 & \text{if } k=1 \\
        \mu_{k-1} + \exp(\theta_k) & \text{if } 1 < k \leq K.
    \end{cases}
\end{aligned}
\end{equation}

\subsection{Case Study 1: Gaussian mixture model}

\begin{figure}[!ht]
    \centering
    \includegraphics[width=0.8\linewidth]{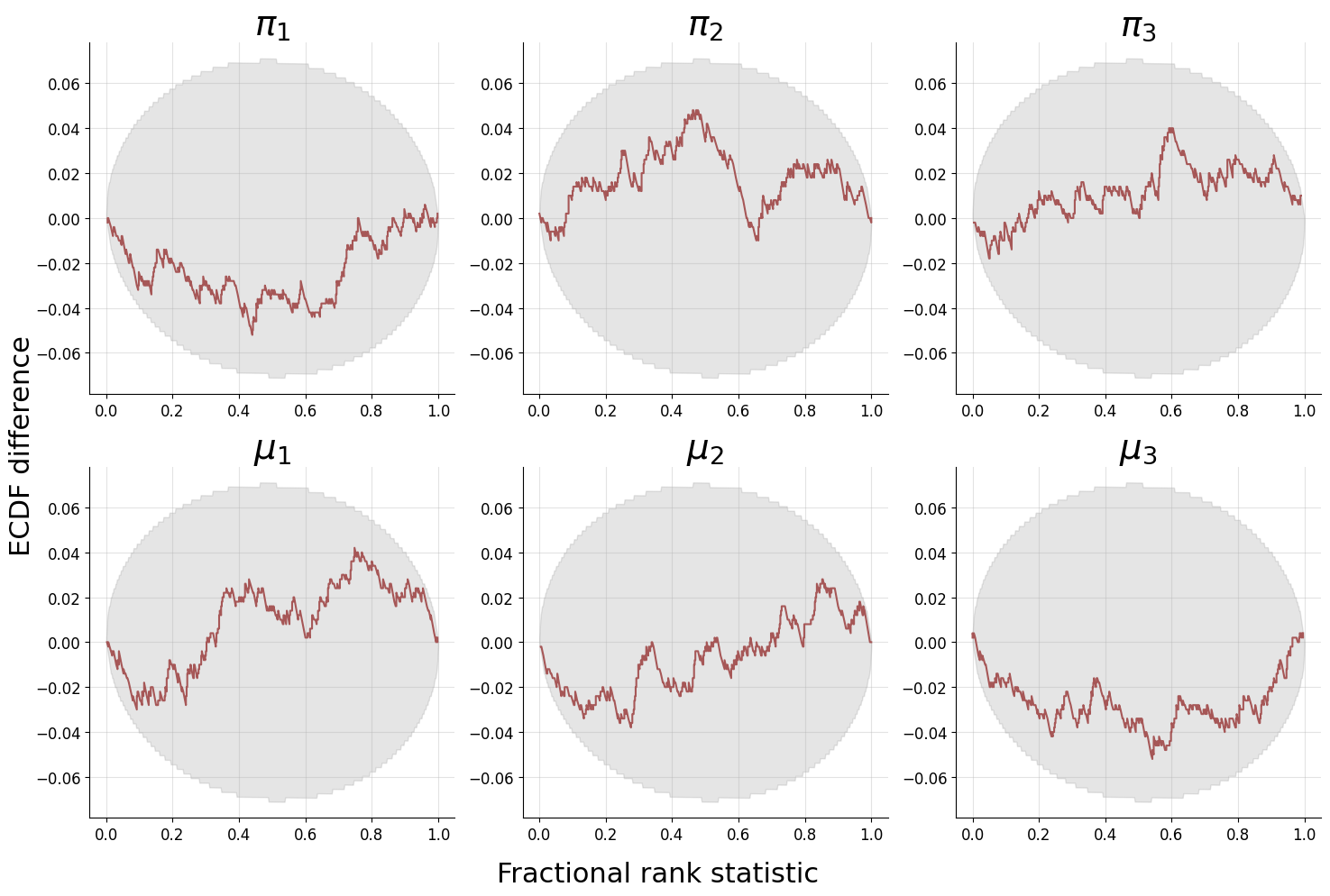}
    \caption{\emph{Case Study 1: Gaussian mixture model.} Simulation-based calibration results displayed as difference between the empirical and expected cumulative distribution function of the fractional rank statistic. The shaded area corresponds to the 95\% Confidence bands.}
    \label{fig:mixture-normal_sbc-ecdf-constrained}
\end{figure}

The first experiment used as an example is a simple independent mixture model with three mixture components. The generative model is as follows:
\begin{equation}
\label{eq:mixture-normal}
    \begin{aligned}
        N & \sim \text{Uniform}(150, 250) \\
        P & \sim \text{Uniform}(2, 4) \\
        (\mu_1, \mu_2, \mu_3) & \sim \text{Normal}\Big((-2, 0, 2), \mathbb{I} \Big)_{\mu_1 < \mu_2 < \mu_3} \\
        \pi & \sim \text{Dirichlet}(2, 2, 2) \\
        z_i & \sim \text{Categorical}(\pi) & \text{for } i \in \{ 1, \dots, N \} \\
        y_{ij} & \sim \text{Normal}(\mu_{z_{i}}, 1) & \text{for } i \in \{ 1, \dots, N \}; j \in \{ 1, \dots, P \}
    \end{aligned}
\end{equation}
The context variables $N$ and $P$ (number of observational units and number of observations per unit, respectively) varied during training so that the networks learn to amortize over different dataset sizes.

The Deep Set architecture \citep{zaheer_deep_2017} produces embedding for exchangeable data, such that the summary embedding is identical for any permutation of the data points. We used a Deep Set network as the local summary network $h_\omega$ to obtain an embedding for each $y_i$ individually, producing a set of embeddings $\{h_\omega(y_i)\}$. This approach compresses the data at the subject level, ensuring that the representation of each observational unit has a consistent dimensionality, regardless of the number of observations per unit $p$.
In this case study, the local summary could be easily handcrafted (i.e., by computing the arithmetic mean). To showcase the capabilities of the framework in general, we still use a neural network, albeit it need not be complex: it contains two dense layers in the inner and outer function of the Deep Set, as well as two equivariant layers. The output size of the summary network was set to two: While in principle the summary network could capture all information with a single output (the minimal number of sufficient statistics is $1$ in this case), giving the network more flexibility allowed it to learn the representation more easily.

The global summary network distills a fixed length embedding $h_\psi(\{h_\omega(y_i)\})$ from the set of individual local summaries, $\{h_\omega(y_i)\}$). The global embedding is used for parameter posterior inference, and so has a more demanding task than the local inference network; if only because it needs to extract summary statistics that relate to the five free parameters in the model. The global network is also implemented as a Deep Set \citep{zaheer_deep_2017}, but we let the network possess more expressive power; the number of dense layers in the inner and outer functions were doubled (i.e., four), while the number of equivariant layers was set to three. 

\begin{figure}[t]
    \centering
    \includegraphics[width=0.8\linewidth]{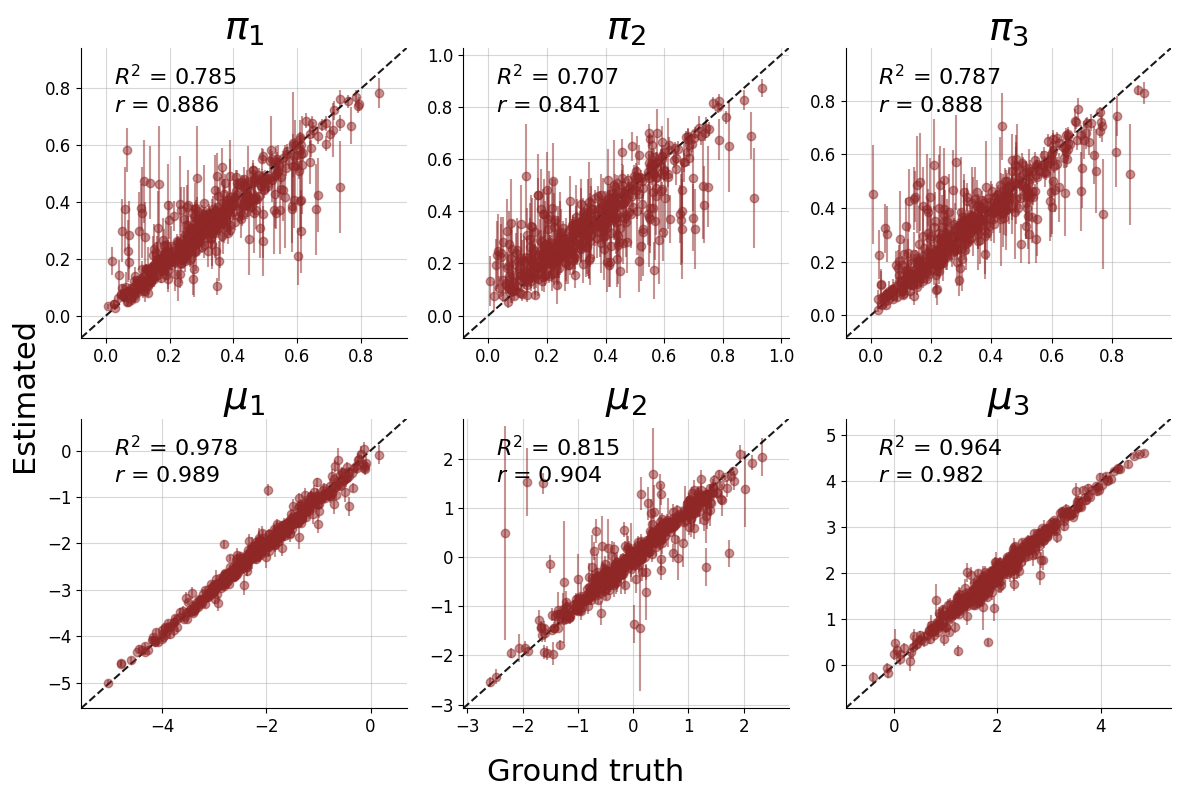}
    \caption{\emph{Case Study 1: Gaussian mixture model.} Parameter recovery shown as scatter plot between the true data generating parameter values and the estimated parameter values. The point estimates are the median, whereas the lines depict the 95\% central credible interval.}
    \label{fig:mixture-normal_sbc-parameter-recovery-constrained}
\end{figure}

The embedding from the global summary network is concatenated with the context variables $n$ and $p$ and passed to an invertible spline coupling network with 14 layers, implemented according to \citet{radev_bayesflow_2022}. This network is used for parameter posterior inference. As a classification network, a multilayer perceptron model composed of 12 fully connected layers with the ReLU activation function was used.

\begin{figure}[!ht]
    \centering
    \includegraphics[width=\linewidth]{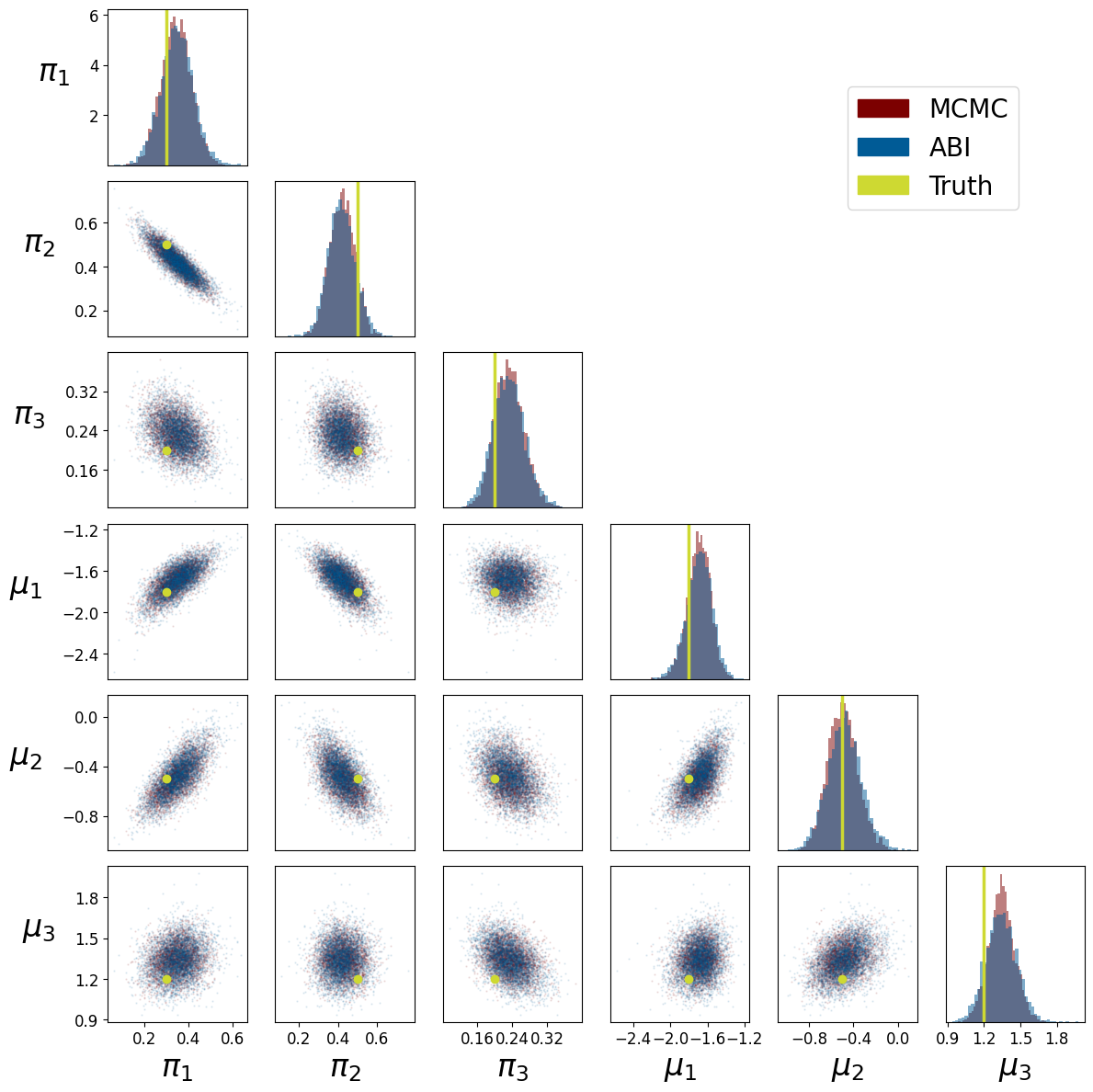}
    \caption{\emph{Case Study 1: Gaussian mixture model.} Joint posterior distribution of the parameters obtained with ABI and MCMC. Example using synthetic data. The diagonals show the marginal parameter posteriors, whereas off-diagonal elements show the pairwise scatter plots to display dependencies between parameters.}
    \label{fig:mixture-normal-joint-posterior}
\end{figure}

All networks were trained jointly in an online training regime, for a total duration of 100 epochs, 2000 iterations each, with each iteration made of 128 sampled data sets from the Bayesian generative model. Figure~\ref{fig:mixture-normal_sbc-ecdf-constrained} shows the results of the simulation-based calibration \citep{talts_validating_2018} of the approximator of the parameter posteriors when the context variables were fixed at $N=200, P=3$. The results indicate good calibration of the estimated posteriors. Parameter recovery plots (Figure~\ref{fig:mixture-normal_sbc-parameter-recovery-constrained}) did not reveal issues with posterior approximation. Obtaining 1000 samples for 500 data sets with \texttt{BayesFlow}, without GPU acceleration, took about 12 sec in total. Obtaining 500 warmup and 1000 sampling iterations from a single chain in \texttt{Stan} (without parallelization) takes about 3 sec on the same hardware for a single data set -- fitting the model on 500 data sets would have therefore taken about a day. Inference with \texttt{BayesFlow} is therefore about 2 orders of magnitudes faster than with \texttt{Stan} for this model.

\begin{figure}[!ht]
    \centering
    \includegraphics[width=\linewidth]{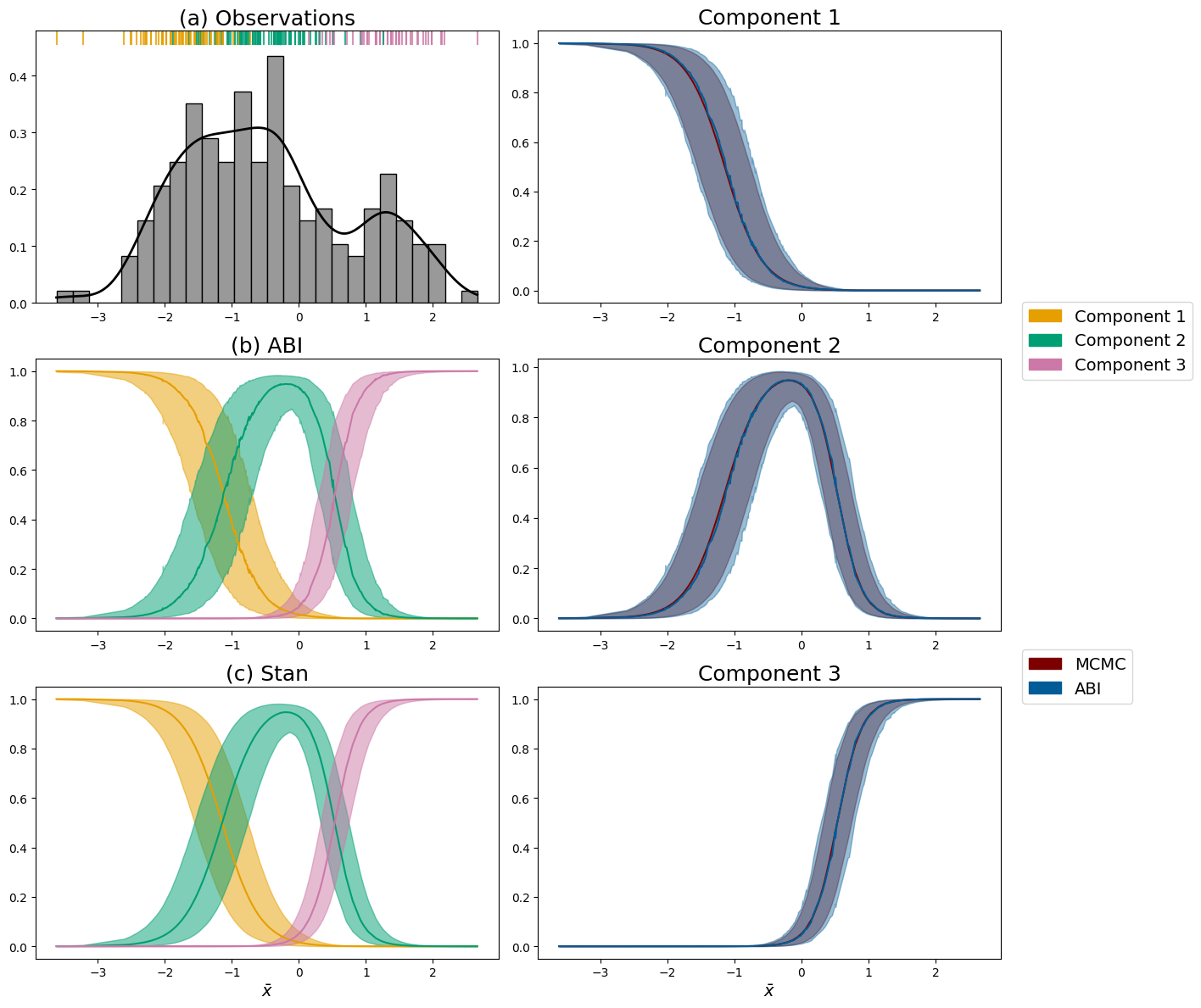}
    \caption{\emph{Case Study 1: Gaussian mixture model.} Comparison of the classification based on the normal mixture model with ABI and MCMC. The data set was generated according to the model in Eq.~\ref{eq:mixture-normal} with $n = 200$ and $p = 3$. The $x$-axis displays the mean for each subject across the three observations. Panel a shows the marginal distribution of the means per observational unit, with rug marks on top showing the individual points coming from three different mixture components. Panel b shows probability of membership in each of the respective mixture components, given the mean on the $x$-axis, according to ABI. Panel c shows the same but using MCMC. Panels on the right side show the same components probabilities, ABI and MCMC estimates overlayed on top of each other to allow direct comparison. The lines show the median of the classification distribution, the confidence bands display the 95\% central credible interval.}
    \label{fig:mixture-normal-classification}
\end{figure}

Next, we show how the amortized inference compares to results obtained using MCMC with \texttt{Stan} \citep{carpenter_stan_2017}. We simulated a single data set from the generative model, and fitted the mixture model on the simulated data with ABI and with MCMC. The true parameter values in this particular example are $\pi = (0.3, 0.5, 0.2)$, $\mu = (-1.8, -0.5, 1.2)$ and the context variables were set to $N=200, P=3$.

Figure~\ref{fig:mixture-normal-joint-posterior} shows the joint posterior distribution of the parameters as estimated by ABI and MCMC, with 6000 samples using either method. Visual inspection indicates that the distributions are almost identical. The C2ST score is 0.53, indicating that distinguishing between the ABI and MCMC samples is a difficult task for a neural classifier. These results suggest that both methods sample from nearly the same distribution.

Figure~\ref{fig:mixture-normal-classification} shows the results of the classification network when applied to a single data set drawn from the Bayesian generative model. The results indicate that the neural estimation of the classification probabilities are similar to the results calculated using the analytic likelihoods based on the parameter posteriors obtained with MCMC. The results suggest that the neural classifier is able to represent the mixture probabilities in a faithful manner. The 95\% CI intervals also overlap for both methods, suggesting that the neural classifier is able to accurately propagate the uncertainty of the parameter values into the uncertainty of the classification probabilities. However, the neural estimates are visibly more ``jittery'', i.e., rather than producing smooth changes in the input to small change in the input, the output varies slightly more chaotically than expected. This might be caused by an overly complex classification network (12 fully connected layers) with a ReLU activation (which is sometimes prone to produce non-smooth output).

\subsection{Case Study 2: Gaussian hidden Markov model}

\begin{figure}
    \centering
    \includegraphics[width=\linewidth]{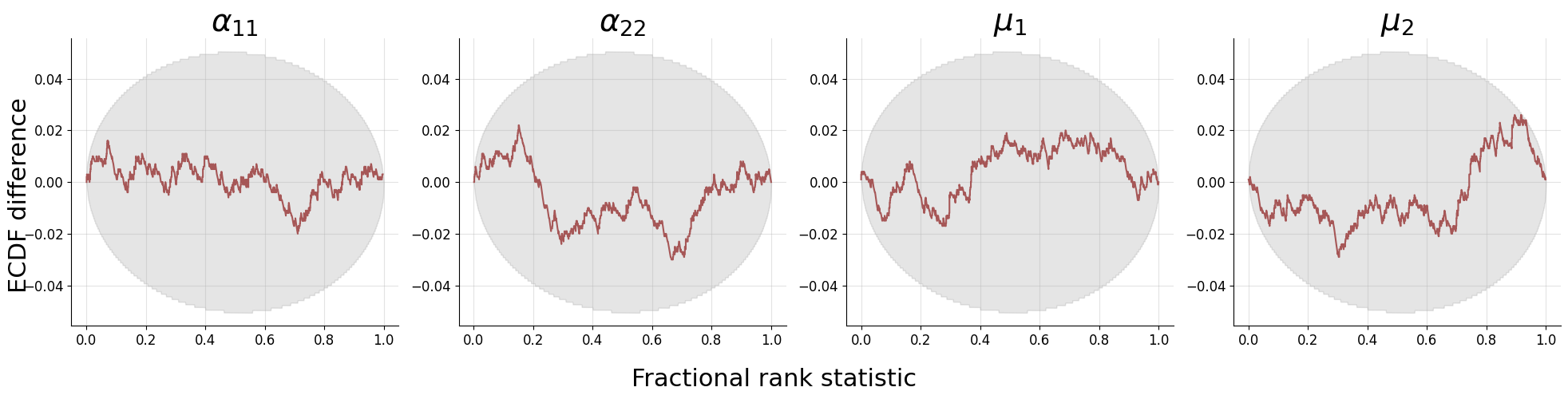}
    \caption{\emph{Case Study 2: Gaussian hidden Markov model.} Simulation-based calibration results displayed as difference between the empirical and expected cumulative distribution function of the fractional rank statistic. The shaded area corresponds to the 95\% Confidence bands.}
    \label{fig:hmm-normal_sbc-diff_constrained}
\end{figure}

The second experiment builds on the first one, but introduces some changes. First, instead of the latent mixture indicators being sampled independently, they now follow a first-order Markov chain. The model can be therefore viewed as a hidden Markov model for evenly spaced time-series \citep{fruhwirth-schnatter_finite_2006,visser_mixture_2022,zucchini_hidden_2016}. Second, each time point can consist of variable number of observations, instead of that being fixed for all time points. Lastly, the model is composed of only two mixture components (hidden states). The model can be written down as follows,
\begin{equation}
\begin{aligned}
P_{i} & \sim \text{Uniform}(2, 5) & \text{for } i \in \{ 1, \dots, N \}\\
\alpha_k & \sim \text{Dirichlet}(2, 2) & \text{for } k \in \{1, 2\} \\
(\mu_1, \mu_2) & \sim \text{Normal}\Big((-1.5, 1.5), \mathbb{I} \Big)_{\mu_1 < \mu_2} \\
z_{1} & \sim \text{Categorical}(0.5, 0.5) \\
z_{i} & \sim \text{Categorical}(\alpha_{z_{i-1}}) & \text{for } i \in \{ 2, \dots, N \}\\
y_{ij} & \sim \text{Normal}(\mu_{z_{i}}, 1) & \text{for } i \in \{ 1, \dots, N \}; j \in \{ 1, \dots, P_i\},\\
\end{aligned}
\end{equation}
where $N=100$ is the number of time points (observational units) and $P_i$ is the number of observations per time point. The $2\times2$ transition matrix with elements $\alpha_{ij}$ gives the probability of transitioning from state $i$ to state $j$. Since each row of the transition matrix sums up to one, we will only show the results for the diagonal elements of the matrix $\alpha_{11}$ and $\alpha_{22}$, corresponding to the probability of staying at the current state 1 or 2, respectively.

For the same reasons as in the first case study, Deep Set architecture \citep{zaheer_deep_2017} was used as the local summary network $h_\omega$. However, in order to extract information from the temporal dependencies between the data points, an LSTM network \citep{gers_learning_1999} was used as the global summary network.

The classification network $f_\alpha$ is implemented as an LSTM layer with 32 units, followed by series of fully connected dense layers with ReLU activation. The LSTM layer allows the network to take into account other observational units, whereas the dense layers increase the expressiveness of the network.

\begin{figure}[!t]
    \centering
    \includegraphics[width=\linewidth]{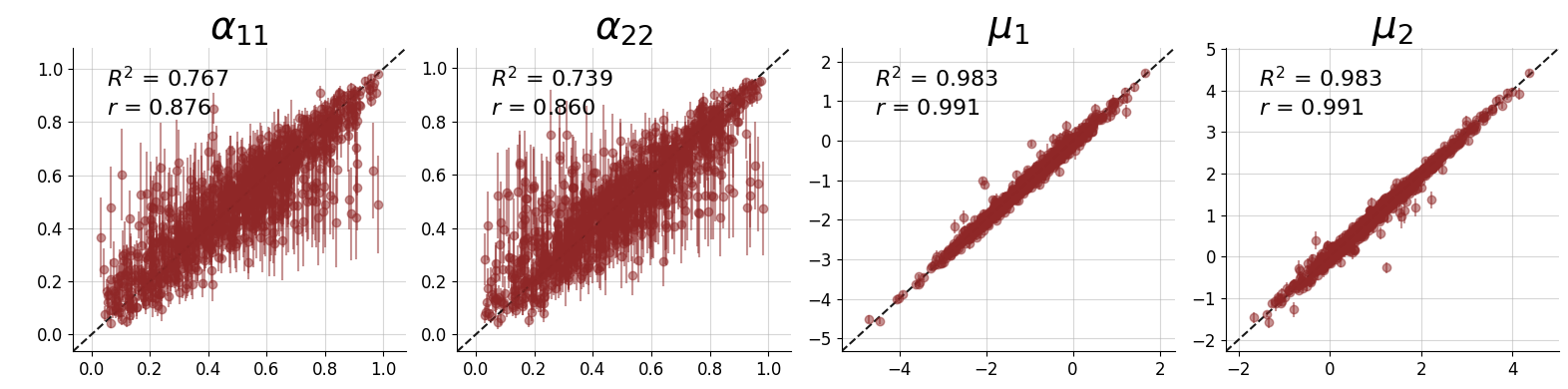}
    \includegraphics[width=\linewidth]{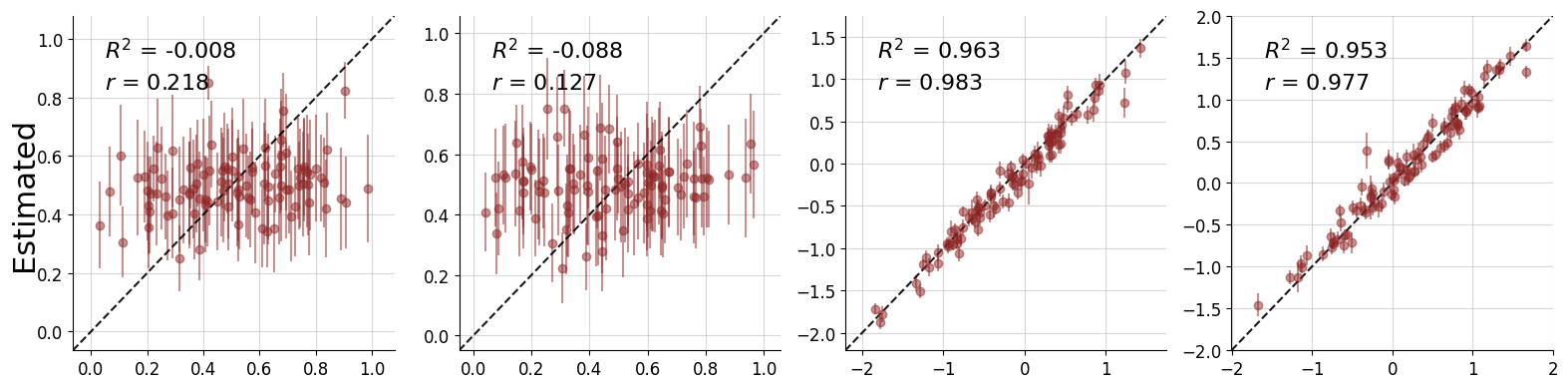}
    \includegraphics[width=\linewidth]{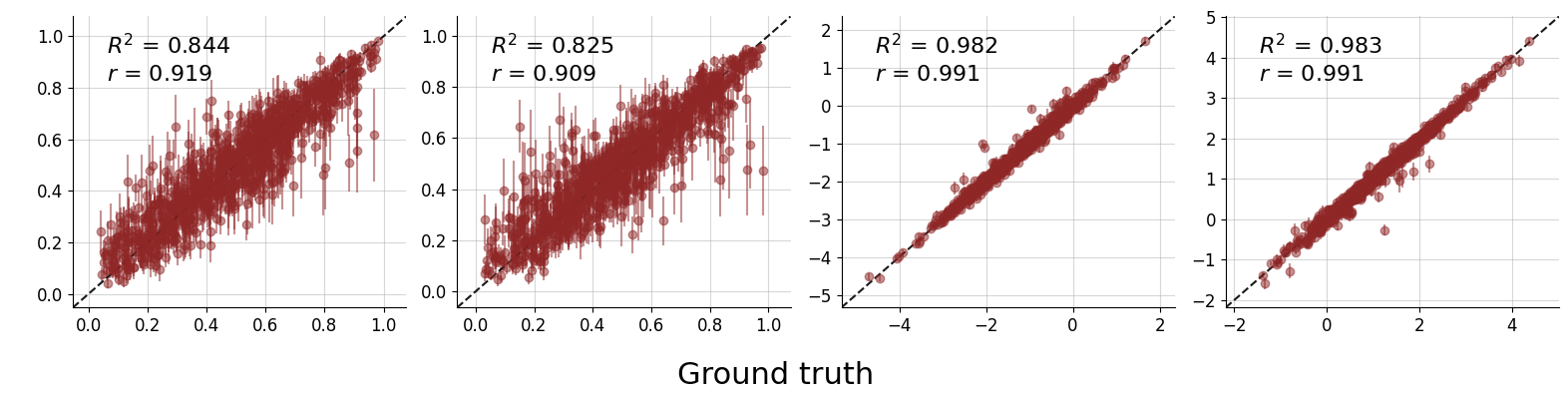}
    \caption{\emph{Case Study 2: Gaussian hidden Markov model.} Parameter recovery shown as scatter plot between the true data generating parameter values and the estimated parameter values. The point estimates are the median, whereas the lines depict the 95\% central credible interval. Top row: Recovery across the entire parameter space. Middle row: Recovery for a subset of the parameter space where $\mu_2 - \mu_1 < 2/3$. Bottom row: Recovery for a subset of the parameter space where $\mu_2 - \mu_1 \geq 2/3$.}
    \label{fig:hmm-normal_recovery}
\end{figure}

All networks were trained jointly in an online training regime, for a total duration of 50 epochs, 500 iterations each, with each iteration made of 512 sampled data sets from the Bayesian generative model. 

For validation of the posterior approximator, we simulated 1000 data sets using the Bayesian generative model, and fitted them with the amortized posterior approximator by generating 1000 posterior samples for each data set. Obtaining 1000 samples for 1000 data sets with \texttt{BayesFlow}, even without GPU acceleration, took about 12 sec. Obtaining 500 warmup and 1000 sampling iterations from a single chain in \texttt{Stan} (without parallelization) takes about 2 sec on the same hardware \emph{for a single data set}. Inference with ABI is therefore about 2 orders of magnitudes faster than with MCMC for this model. As shown in Figure~\ref{fig:hmm-normal_sbc-diff_constrained}, the SBC revealed no clear patterns of miscalibration of the posterior approximator. Figure~\ref{fig:hmm-normal_recovery} does not reveal issues with estimating the mean parameters $\mu_1$ and $\mu_2$. Recovery of the transition probabilities ($\alpha_{11}$ and $\alpha_{22}$) is slightly worse. A closer inspection reveals that parameter recovery is challenging in regions of the parameter space where the latent states are insufficiently separated ($\mu_2 - \mu_1 \geq 2/3$), as illustrated in the middle row of Figure~\ref{fig:hmm-normal_recovery}. In contrast, when the states are well separated, recovery of the transition probabilities improves (bottom row of Figure~\ref{fig:hmm-normal_recovery}). This pattern is characteristic of mixture models: poor separation between components increases uncertainty in the mixture indicators, which in turn reduces the precision of estimates for mixture or transition probabilities.

\begin{figure}[!t]
    \centering
    \includegraphics[width=\linewidth]{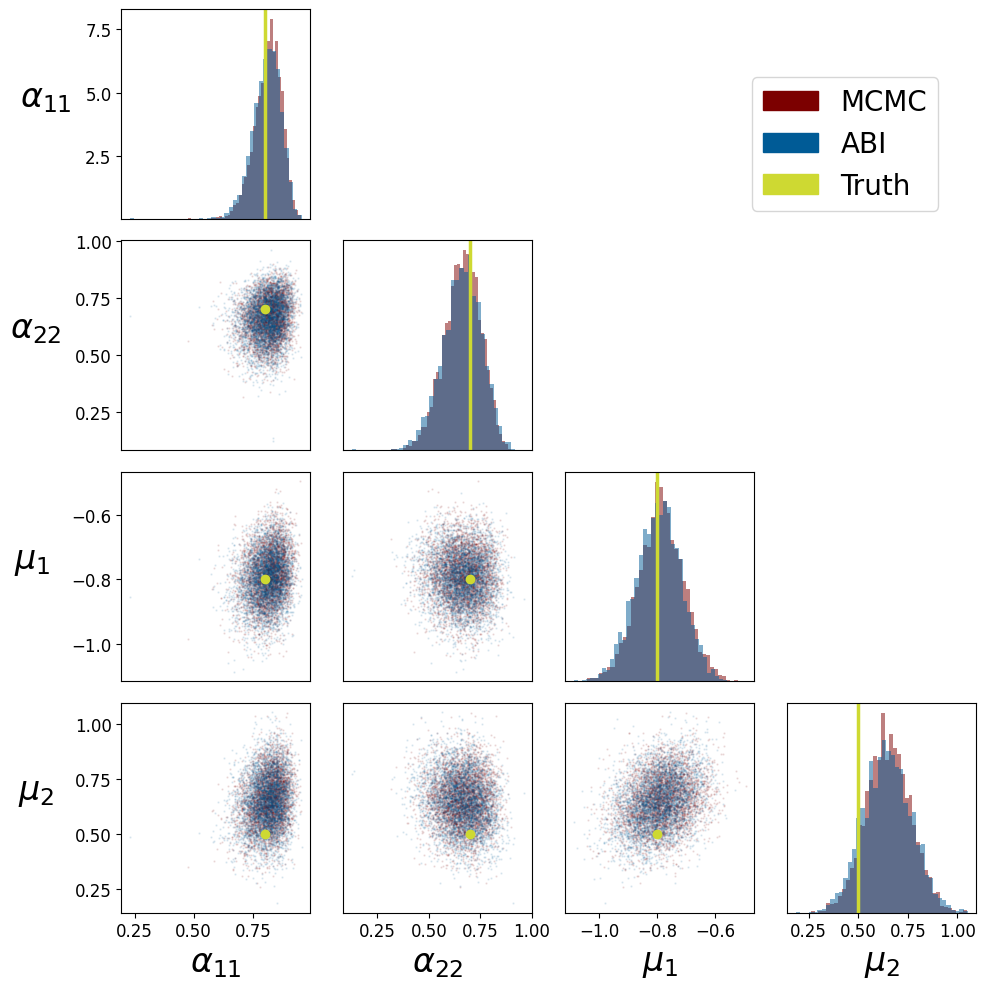}
    \caption{\emph{Case Study 2: Gaussian hidden Markov model.} Comparison between parameter posterior approximations using ABI and MCMC. Example using synthetic data. The diagonals show the marginal parameter posteriors, whereas off-diagonal elements show the pairwise scatter plots to display dependencies between parameters. The two methods return nearly identical results, yielding almost perfectly overlapping distributions.}
    \label{fig:hmm-normal_joint-posterior}
\end{figure}

To showcase the application of the amortized HMM, we show the results applied to one synthetic example generated from the Bayesian generative model. Figure~\ref{fig:hmm-normal_joint-posterior} shows the posterior distribution of the parameters. Both ABI and MCMC estimates are very similar. The C2ST score is 0.56, indicating that distinguishing between the ABI and MCMC samples is a difficult task for a neural classifier. These results suggest that both methods sample from nearly the same distribution.

\begin{figure}[!t]
    \centering
    \includegraphics[width=\linewidth]{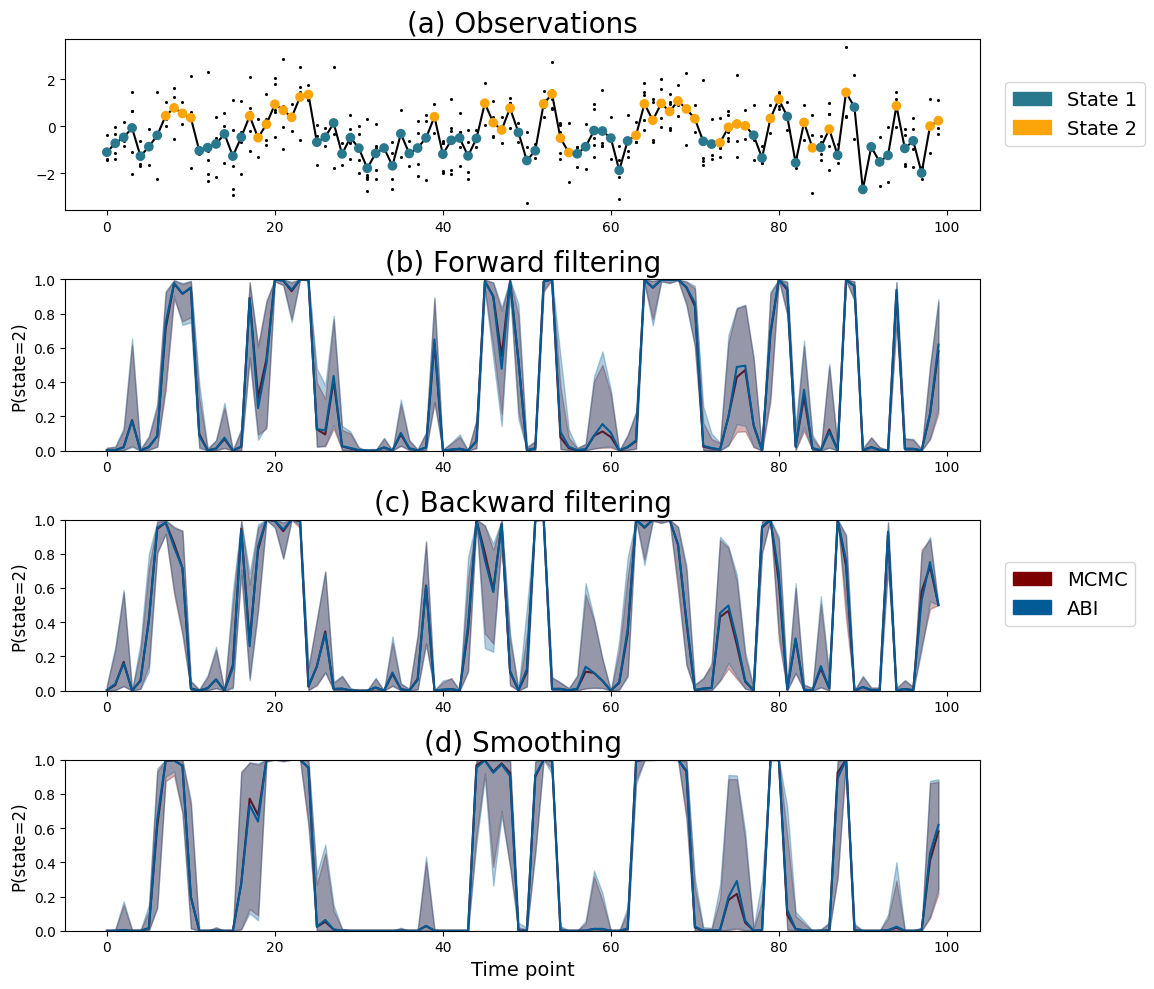}
    \caption{\emph{Case Study 2: Gaussian hidden Markov model.} Comparison between ABI and MCMC in the predicted classification probabilities using forward (panel b) and backward (panel c) filtering, and smoothing (panel c), using the Gaussian hidden Markov model. In panel a, individual observations are shown as black dots. Large colored dots depict the mean. In the classification plots, the solid lines show the posterior median, and the confidence bands display the 99\% central credible interval. Example using synthetic data. The two methods return nearly identical results, yielding almost perfectly overlapping classification probabilities.}
    \label{fig:hmm-normal_classification}
\end{figure}

ABI and MCMC also agree on the classification probabilities based on forward filtering, backward filtering, and smoothing, as shown in Figure~\ref{fig:hmm-normal_classification}, suggesting that the classification network is well calibrated as well, including calibration of the classification uncertainty as demonstrated by overlapping confidence intervals.

\subsection{Case Study 3: Latent switches in cognitive processing}

\begin{figure}
    \centering
    \includegraphics[width=\linewidth]{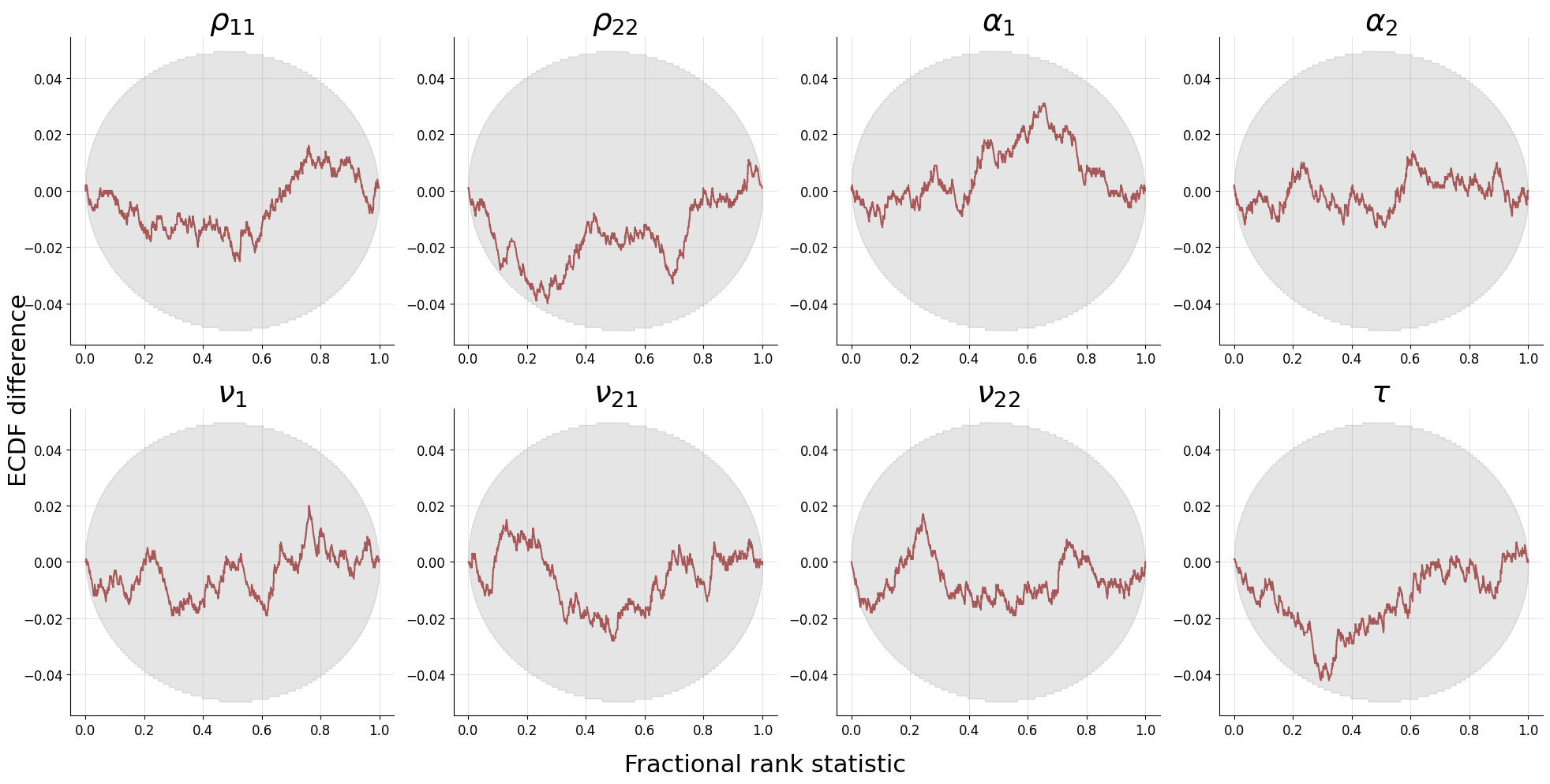}
    \caption{\emph{Case Study 3: Latent switches in cognitive processing.} Simulation-based calibration results displayed as difference between the empirical and expected cumulative distribution function of the fractional rank statistic. The shaded area corresponds to the 95\% Confidence bands.}
    \label{fig:hmm-eam-sbc-ecdf-diff-constrained}
\end{figure}

The current application is based on the empirical study reported by \citet{dutilh_phase_2011} that studied human decision making in speeded decision tasks. One prediction is that under different incentive conditions (i.e., varying reward for speed versus accuracy), participants are unable to control the speed-decision trade-off on a continuum, but rather switch between distinct modes of behavior -- under one mode, participants tend to guess randomly in order to provide fast responses, and under another mode, provide slower responses for the sake of improving their accuracy. These modes can be represented as latent states in a mixture model. Since the incentives in the experiment between trials are adjusted continually, it is expected that there will be some temporal dependencies between the states; here, we will model these dependencies with a Hidden Markov model.

The response behavior is modeled using evidence accumulation models (EAMs), a widely used class of cognitive process models of decision-making \citep[e.g.,][]{ratcliff1978theory,ratcliff2008diffusion,bogacz2006physics,ratcliff2016diffusion,evans2020evidence}. EAMs assume that a decision is made by stochastically accumulating evidence over time toward a boundary that represents the amount of evidence required to trigger a response. The drift rate ($\nu$) governs the mean rate of accumulation, while the boundary separation ($\alpha$) reflects response caution: larger boundaries lead to slower but more accurate responses. A non-decision time ($\tau$) accounts for residual processes such as perceptual encoding and motor execution. Noise in evidence accumulation can arise both within and between trials, reflecting sensory noise or fluctuations in attention and task difficulty, etc \citep{tillman_sequential_2020,bogacz2006physics,evans2020evidence}.

Under the \emph{guessing state} ($z_i=1$), response times arise as a result of a simple evidence accumulation process as a Wiener diffusion process \citep[see][]{karatzas2014brownian} with a drift $\nu_1$. Once evidence reaches the threshold $\alpha_1$, a response is triggered at random with a non-decision time delay $\tau$. This response time process is known as simple response times \citep{luce_response_1991}; using the Wiener diffusion process implies that the response times follow the shifted Wald distribution \citep[i.e., inverse Gaussian;][]{anders_shifted_2016,chhikara_inverse_2019}. 

Under the \emph{controlled state} ($z_i = 2$), responses are generated from a four-parameter Racing Diffusion Model \citep[RDM,][]{tillman_sequential_2020}. 
The RDM assumes that two parallel accumulators (one representing the correct response and one the incorrect response) race towards a decision boundary $\alpha_{22}$. Each accumulator follows a Wiener diffusion process, characterized by its own drift rate ($\nu_{22}$ for the correct accumulator and $\nu_{21}$ for the incorrect accumulator). A decision is made when the first accumulator reaches the boundary $\alpha_2$, determining both the response outcome $y_i$ (1 if the ``correct'' accumulator wins, 0 otherwise) and the decision time $t_i$. A non-decision time $\tau$ is then added to account for sensory encoding and motor execution, such that the observed response time is $rt_i = t_i + \tau$.

The RDM captures characteristic phenomena of human decision-making \citep{tillman_sequential_2020}, including the continuous speed--accuracy trade-off: increasing the boundary $\alpha_2$ or the drift rate $\nu_{22}$ leads to slower but more accurate responses. Despite its cognitive richness, the four-parameter RDM remains relatively parsimonious, making it a suitable observation model within a larger hierarchical framework \citep[in this case, a single state of the HMM;][]{kucharsky_hidden_2021}.

The full model can be summarized as follows, 
\begin{equation}
\begin{aligned}
\rho_k & \sim \text{Dirichlet}(2, 2) & \text{for } k \in \{1, 2\} \\
z_{1} & \sim \text{Categorical}(0.5, 0.5) \\
z_{i} & \sim \text{Categorical}(\rho
_{z_{i-1}}) & \text{for } i \in \{ 2, \dots, N \}\\
\alpha_1 & \sim \text{Normal}(0.5, 0.3)_{T(0, \infty)} \\ 
\nu_1 & \sim \text{Normal}(5.5, 1.0)_{T(0, \infty)} \\ 
(\alpha_2 - \alpha_1) & \sim \text{Normal}(1.5, 0.5)_{T(0, \infty)} \\
\nu_{21} & \sim \text{Normal}(2.5, 0.5)_{T(0, \infty)} \\
(\nu_{22} - \nu_{21}) & \sim \text{Normal}(2.5, 1.0)_{T(0, \infty)} \\
\tau & \sim \text{Exponential}(5.0) \\
(rt_{i}, y_i) & \sim \begin{cases} \big(\text{Wald}(\alpha_1, \nu_1) + \tau, \text{Bernoulli}(0.5)\big) & \text{if } z_{i} = 1\\
\text{RDM}(\alpha_2, \nu_{21}, \nu_{22}, \tau) & \text{if } z_{i} = 2
\end{cases} & \text{for } i \in \{ 1, \dots, N \},
\end{aligned}
\end{equation}
where $N=400$ is the number of trials in the experiment (observational units). The observable variables are the response times (in seconds) $rt_i$ and the choices (binary) $y_i$ on experimental trial $i$. The $2\times2$ transition matrix with elements $\rho_{ij}$ gives the probability of transitioning from the latent state $i$ to $j$. Since each row of the transition matrix sums up to one, we will only show the results for the diagonal elements of the matrix $\alpha_{11}$ and $\alpha_{22}$, the probability of staying in the guessing and controlled state, respectively. The response model comprises of six parameters: the non-decision time $\tau$, the decision boundary $\alpha_1$ and the drift rate $\nu_1$, respectively, under the guessing state, the decision boundary $\alpha_2$, the drift rate for the incorrect response $\nu_{21}$ and the correct response $\nu_{22}$ under the controlled state. The parameter priors were selected through prior predictive simulations such that model generates realistic patterns of observed data under each state \citep{kucharsky_hidden_2021}.

\begin{figure}[!t]
    \centering
    \includegraphics[width=\linewidth]{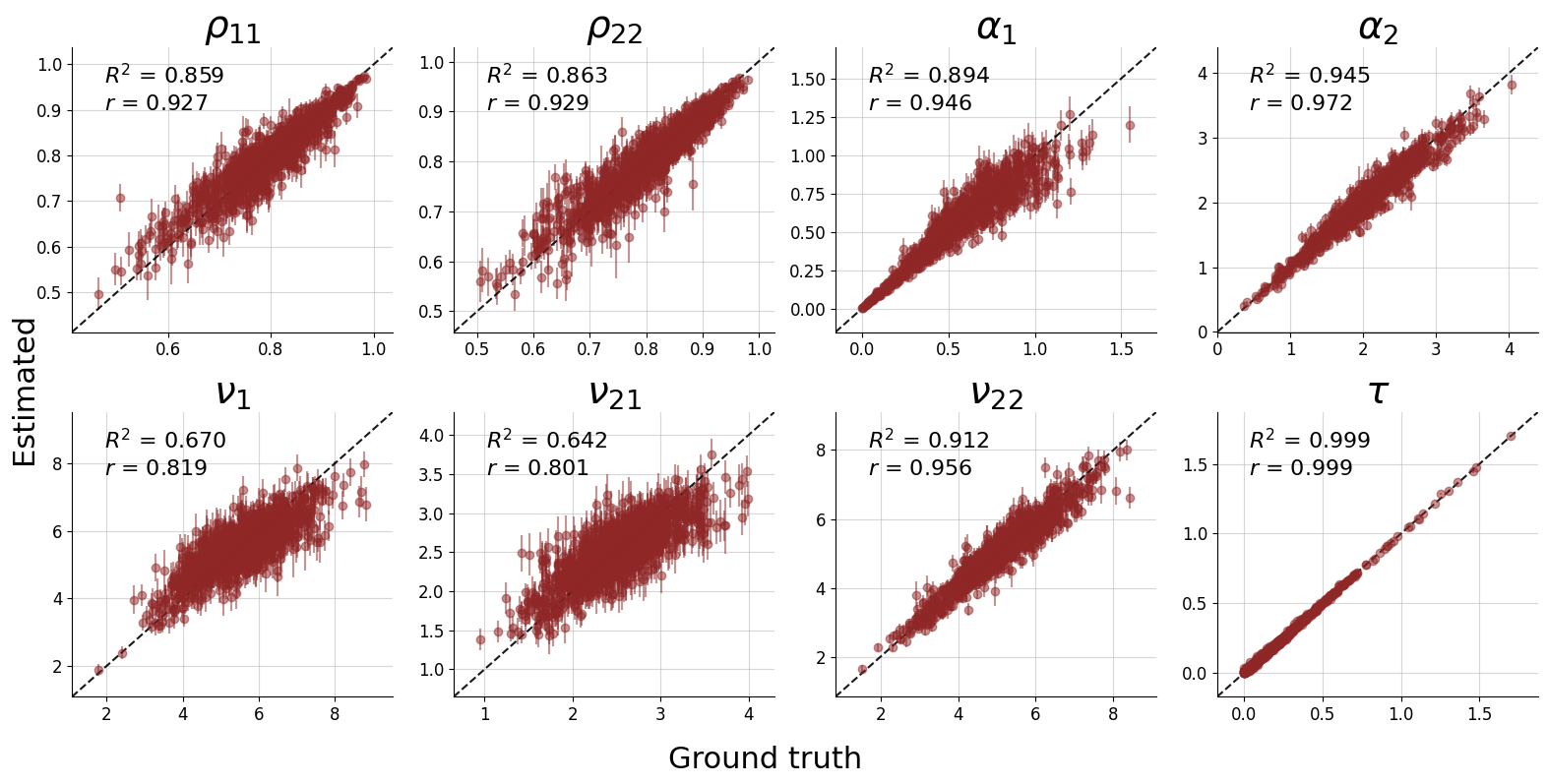}
    \caption{\emph{Case Study 3: Latent switches in cognitive processing.} Parameter recovery shown as scatter plot between the true data generating parameter values and the estimated parameter values. The point estimates are the median, whereas the lines depict the 95\% central credible interval.}
    \label{fig:hmm-eam-recovery-constrained}
\end{figure}

Since the observational units are always comprised only of one observation of response time $rt_i$ and one observation of choice $y_i$, using a local summary network would be redundant. Thus, raw data is passed directly to the global summary network for posterior inference, as well as the classification network for the classification inference. The global summary network consists of four convolutional layers followed by a bidirectional LSTM with 256 units. The convolutional layers serve to extract and smooth local temporal patterns in the sequence of responses, while the LSTM captures short- and medium-range dependencies that reflect latent state persistence and switching in the underlying HMM. This combination allows the inference network to construct summary representations that are sensitive to temporal structure in the observed data. The final output layer (32 units) is regularized using the MMD loss to ensure approximately normal summary statistics \citep{kothe_detecting_2024}. This will allow us later to compute diagnostics of model misspecification when the model is applied to empirical data.

The classification network $f_\alpha$ is implemented as an LSTM layer with 32 units, followed by a series of fully connected layers with ReLU activation. The LSTM layer allows the network to take the temporal dependencies in the data, while the dense layers further increase the expressiveness of the network. The classification network was trained both in the forward and backward regime, so that it can be used for both \emph{filtering} and \emph{smoothing}.

All networks were trained jointly in an online training regime, for a total duration of 100 epochs, 1000 iterations each epoch, with each iteration made of 256 sampled data sets from the Bayesian generative model.

For validating the posterior approximator, we simulated 1000 data sets from the Bayesian generative model, and fitted them with the amortized posterior approximator, obtaining 1000 posterior samples for each data set. As shown in Figure~\ref{fig:hmm-eam-sbc-ecdf-diff-constrained}, SBC revealed no miscalibration of the posterior approximator. Figure~\ref{fig:hmm-eam-recovery-constrained} shows that parameters can also be recovered. Overall the results of the validation simulations suggested that the amortized posterior works well.

To evaluate the model's performance on real data, the model was fitted to data from 11 participants in an experiment reported by \citet{dutilh_phase_2011}. For reference, we used \texttt{BayesFlow} and \texttt{Stan} to fit all data sets with ABI and MCMC, respectively. Obtaining 4000 samples for all 11 data sets took about 40 sec using \texttt{BayesFlow}, and about 15 minutes with \texttt{Stan}.

To check for model misspecification, we computed the MMD of the summary statistics returned by the global summary network, and compared it to the distribution of MMD statistics computed on prior predictive data sets. As shown in Table~\ref{tab:hmm-eam_summaries}, for seven out of the eleven data sets, the computed MMD did not exceed critical values that would identify model misspecification. The C2ST score \citep{lopez2016revisiting} generally agreed with MMD; datasets associated with high MMD are also ones where classifying posterior samples from MCMC and ABI is relatively easier.

\begin{table}
    \centering
    \begin{tabular}{lrrrr}
        \toprule
        \multicolumn{1}{c}{} & \multicolumn{2}{c}{MMD test} \\
        \cline{2-3}
        Data set & Statistic & \textit{p}-value & Pareto $\hat{k}$ & C2ST\\
        \midrule
        A & 3.84 & \textbf{.025} & \textbf{3.81} & 0.99 \\
        B & 3.75 & \textbf{.084} & 0.49 & 0.62 \\
        C & 3.52 & .664 & 0.46 & 0.56\\
        D & 3.68 & .168 & 0.41 & 0.71 \\
        E & 3.67 & .192 & 0.53 & 0.65 \\
        F & 3.44 & .895 & 0.59 & 0.58 \\
        G & 3.77 & \textbf{.072} & \textbf{1.61} & 0.98 \\
        H & 3.67 & .197 & 0.67 & 0.55 \\
        I & 3.63 & .282 & 0.32 & 0.55 \\
        J & 3.72 & .453 & 0.66 & 0.56 \\
        K & 3.78 & \textbf{.060} & \textbf{0.76} & 0.82 \\
        \bottomrule
    \end{tabular}
    \caption{MMD test and Pareto $\hat{k}$ diagnostic. Highlighted values indicate failed diagnostics. Failed MMD test ($\alpha=0.1$) indicates a simulation gap and suggests that the amortized posterior may have a problem generalizing to the current data set and the posterior is therefore not trustworthy. Pareto $\hat{k} > 0.7$ usually indicates that Pareto smoothed importance sampling will not be effective in correcting the amortized posteriors.}
    \label{tab:hmm-eam_summaries}
\end{table}

\begin{figure}
    \centering
    \includegraphics[width=1.0\linewidth]{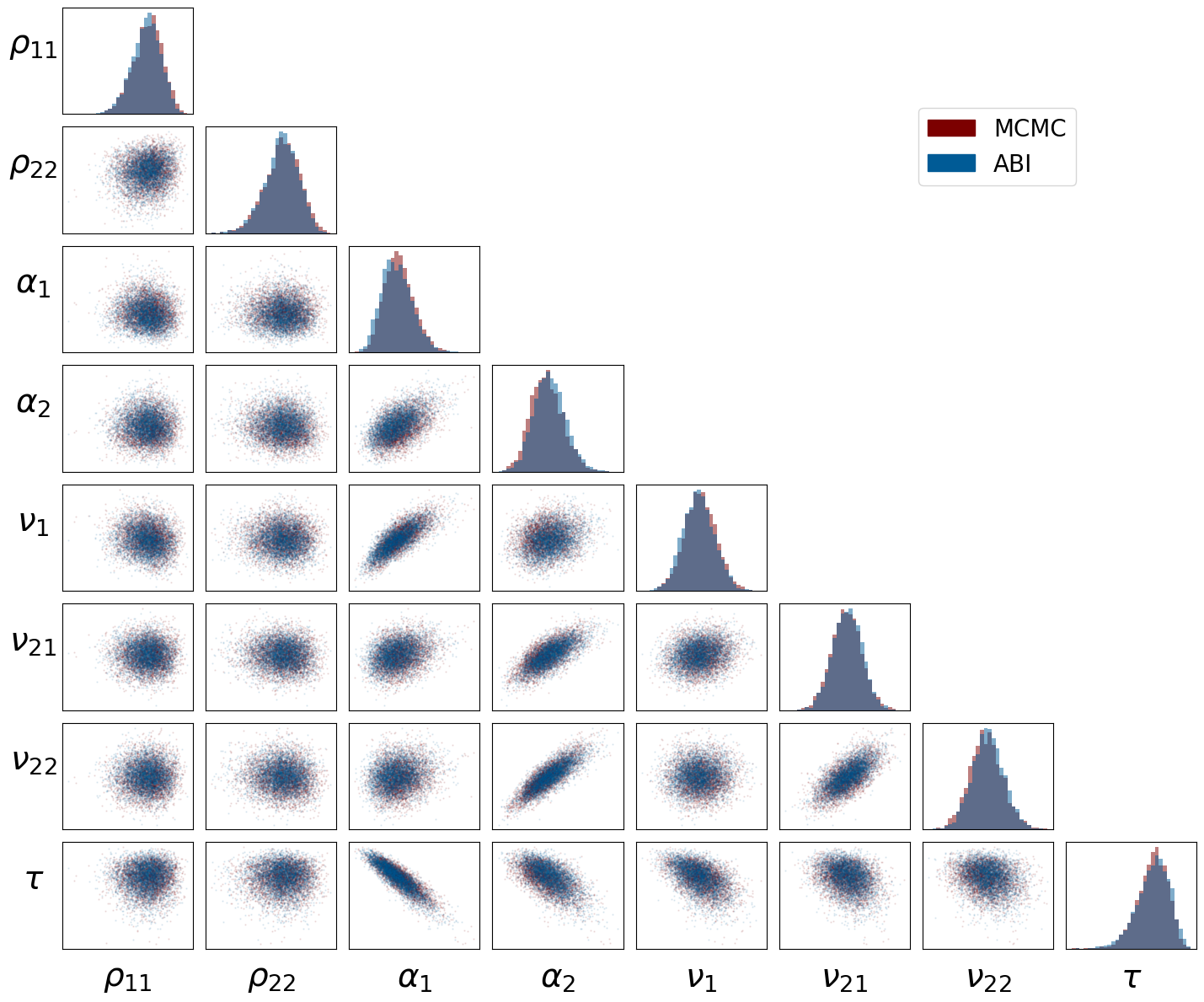}
    \caption{\emph{Case Study 3: Latent switches in cognitive processing.} 
 Comparison of the parameter posterior samples between ABI and MCMC for participant C reported by \citet{dutilh_phase_2011}. The diagonals show the marginal parameter posteriors, whereas off-diagonal elements show the pairwise scatter plots to display dependencies between parameters. The two methods return nearly identical results, yielding almost perfectly overlapping distributions.}
    \label{fig:joint-posterior-subject-c}
\end{figure}

An example of a posterior associated with a typical value of MMD is shown in Figure~\ref{fig:joint-posterior-subject-c}. The posterior approximations of ABI and MCMC almost perfectly overlap, both in terms of their marginal distributions, as well as the join distribution as inspected by pairwise scatter plots. This suggests that the ABI posterior estimate is able to correctly capture the values of the parameters as well as their correlations. The low value of MMD correctly identified that the amortized posterior is indeed well calibrated for this data set, and the estimate is therefore trustworthy.

\begin{figure}
    \centering
    \includegraphics[width=1.0\linewidth]{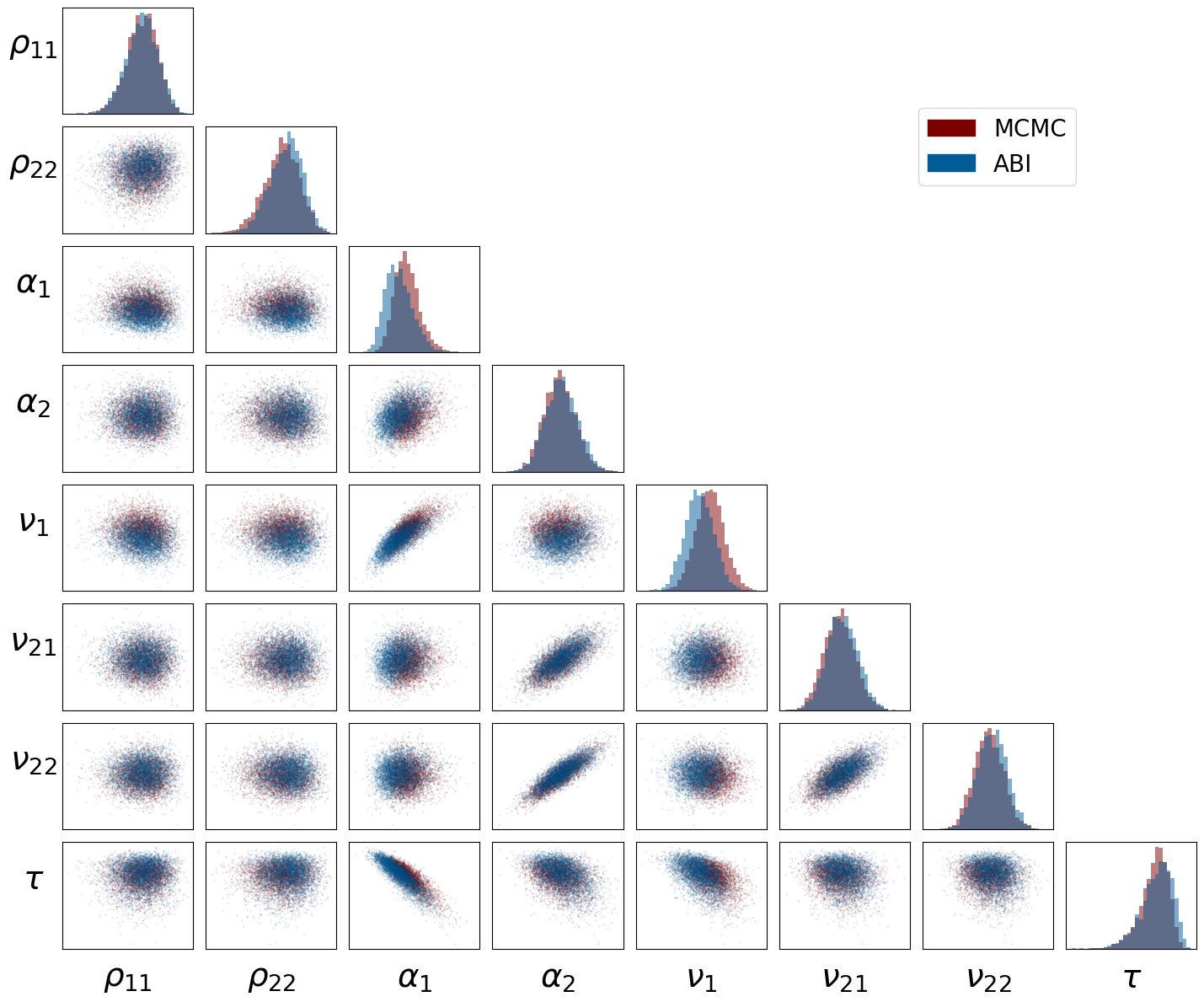}
    \caption{\emph{Case Study 3: Latent switches in cognitive processing.} Comparison of the parameter priors and posterior samples between ABI and MCMC for participant B reported by \citet{dutilh_phase_2011}. The diagonals show the marginal parameter posteriors, whereas off-diagonal elements show the pairwise scatter plots to display correlations between parameters. The ABI approximation is slightly biased but the estimates can be corrected by Pareto smoothed importance sampling.}
    \label{fig:hmm-eam_joint-posterior_subject-b}
\end{figure}

For the other four data sets, the parameter posteriors approximated by ABI may not be trustworthy. For example, as shown for participant B in Figure~\ref{fig:hmm-eam_joint-posterior_subject-b}, it appears that ABI slightly underestimates parameters $\alpha_1$ and $\nu_1$.

Following the amortized Bayesian workflow \citep{schmitt_amortized_2024}, it is possible to correct the approximated posteriors using Pareto smoothed importance sampling \citep[PSIS,][]{vehtari_pareto_2024,vehtari_practical_2017}. For each posterior sample obtained from the neural approximator, we obtain the log posterior density using the neural approximator, as well as the (unnormalized) analytic log posterior density; since we have already implemented this model in \texttt{Stan}, we used the \texttt{Stan} model to compute the analytic density. The log importance weights are calculated as the difference between the log neural density and log analytic density. The Pareto smoothed weights and $k$-statistic were computed using the \texttt{Arviz} package \citep{arviz_2019}.

Table~\ref{tab:hmm-eam_summaries} also shows the Pareto $\hat{k}$ diagnostic. For the data set B, the importance weights can be used for correcting the posterior distribution obtained with ABI.

\begin{figure}
    \centering
    \includegraphics[width=1.0\linewidth]{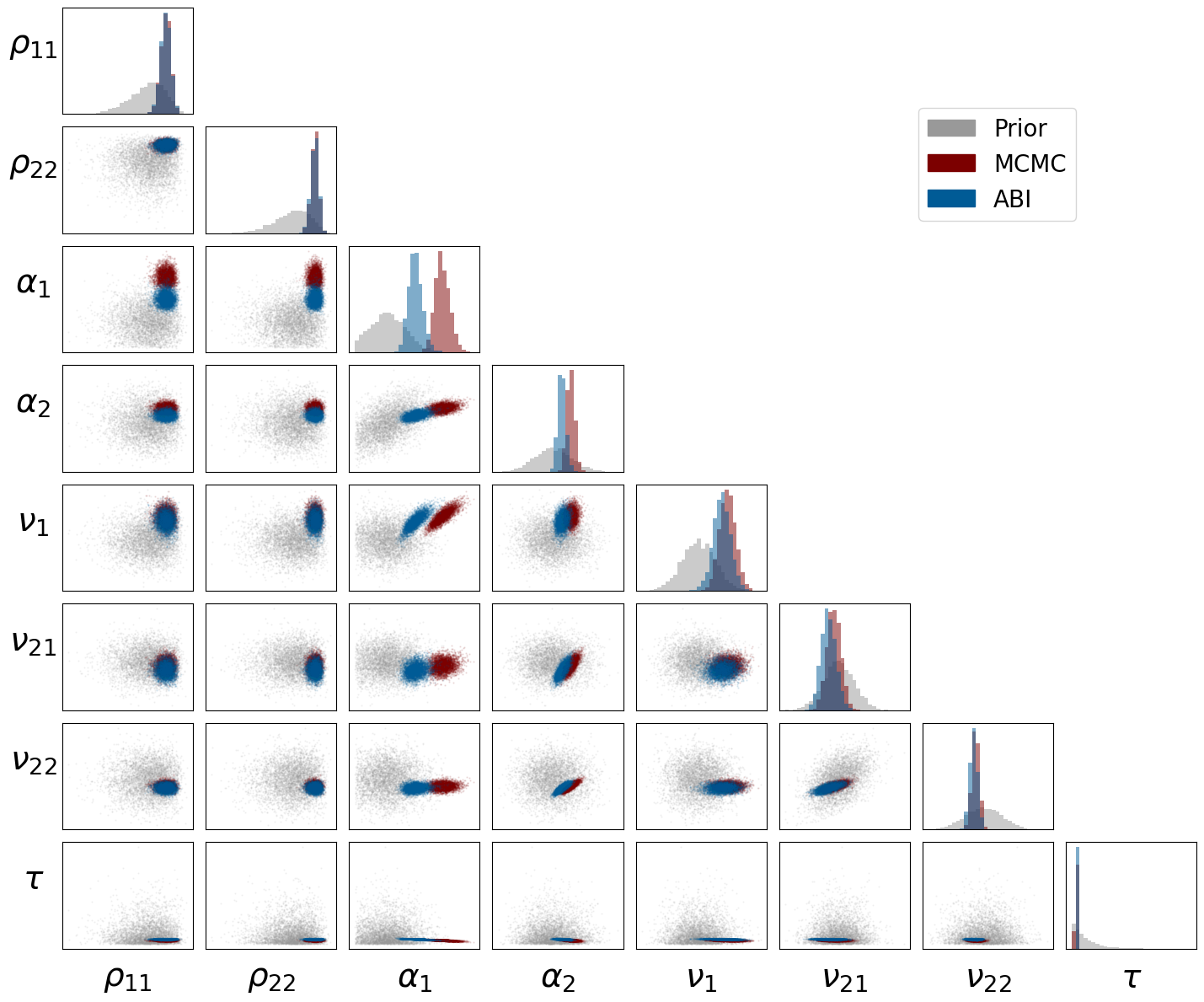}
    \caption{\emph{Case Study 3: Latent switches in cognitive processing.} Comparison of the parameter priors and posterior samples between ABI and MCMC for participant A reported by \citet{dutilh_phase_2011}. The diagonals show the marginal parameter posteriors, whereas off-diagonal elements show the pairwise scatter plots to display dependencies between parameters. The ABI approximation is severely biased and corrections using Pareto smoothed importance sampling will not be efficient.}
    \label{fig:hmm-eam_joint-posterior_subject-a}
\end{figure}

However, for some data sets (mainly, A and G), the ABI posteriors are too far from the true posterior to be corrected by importance sampling. For example, for data set A, the MCMC posterior lies on the tails of the prior for the parameters $\alpha_1$ and $\nu_1$ (Figure~\ref{fig:hmm-eam_joint-posterior_subject-a}). Subsequently, ABI poorly generalizes and substantially underestimates the parameters. This poor generalization is captured by the high MMD value and its \textit{p}-value. Furthermore, the ABI posterior is too far from the true posterior, rendering importance sampling inefficient and unreliable -- which is captured by the high Pareto $\hat{k}$ diagnostic in Table~\ref{tab:hmm-eam_summaries}. This indicates that precise posterior inference for the data sets A and G is not available using the current networks. Data set K is a borderline case with marginally high $\hat{k} = 0.76$. For this data set, importance sampling will likely still work well given enough posterior samples.

\begin{figure}
    \centering
    \includegraphics[width=\linewidth]{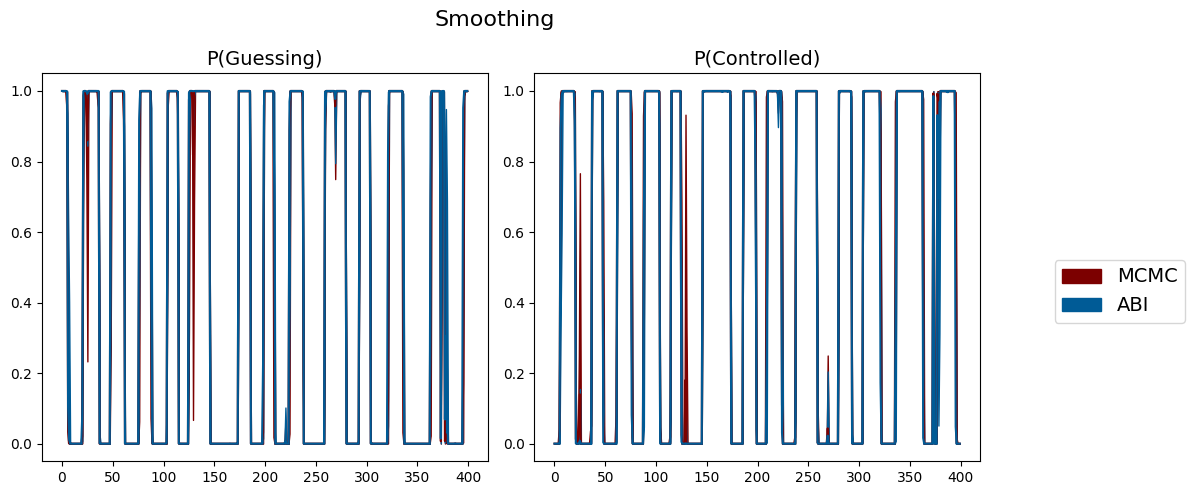}
    \caption{\emph{Case Study 3: Latent switches in cognitive processing.} Comparison between smoothing classification probabilities between ABI and MCMC for participant A reported by \citet{dutilh_phase_2011}. The two methods return nearly identical results, yielding almost perfectly overlapping lines.}
    \label{fig:hmm-eam_smoothing_subject-a}
\end{figure}

Even in the case where the parameter posterior is not estimated reliably using neural networks, the two states are separated well enough that even the biased parameter posteriors do not have a big influence on the resulting classification; as shown in Figure~\ref{fig:hmm-eam_smoothing_subject-a}.

\section{Discussion}
\label{sec:conclusion}

In this paper, we developed and evaluated a framework for amortized Bayesian mixture models based on deep learning. By decomposing mixture models into parameter posterior distributions and mixture membership distributions, we can represent these distributions through corresponding neural architectures. These neural networks can be trained simultaneously on the same training data generated from the Bayesian model. This allows flexible amortization over different design factors, such as the number of observational units. Once trained, the neural networks provide reliable estimates in a fraction of the time required by traditional methods such as MCMC.

We evaluated the proposed framework through three case studies, demonstrating its applicability. The computational faithfulness of the neural networks was rigorously assessed through simulations. We also extensively compared the accuracy of the neural network outputs against state-of-the-art MCMC results obtained using \texttt{Stan} \citep{carpenter_stan_2017}. The first two case studies show a proof of concept on independent and dependent mixture models, though with limited applicability considering their simplistic nature. The third case study brings forth a more realistic scenario of using the framework in empirical context. 

The methodology is implemented with the Python library \texttt{BayesFlow} \citep{radev_bayesflow_2022}, and is publicly accessible online at \href{https://osf.io/7wvyk/}{osf.io/7wvyk/}. 

\subsection*{Limitations \& Future directions}

Although the scope of the statistical models showcased in this article is relatively limited, we believe it demonstrates that the proposed framework has much broader potential. General benefits of ABI, speed during inference, fitting models with intractable likelihoods, might not be the only appealing characteristics in the context of mixture models. For example, some MCMC samplers might be prone to get stuck in isolated parameter regions for specific models \citep{stephens1997bayesian,swanson2024blocked,celeux_computational_2000,diebolt_estimation_1994,marin_bayesian_2005}. ABI might be a promising alternative to MCMC in such use cases.

Due to our aim to validate our results against MCMC, the parametrizations used in this article are inspired by standard practices in probabilistic programming languages such as \texttt{Stan} \citep{carpenter_stan_2017}. However, translating these directly for ABI is not always straightforward. Priors convenient in \texttt{Stan} in particular or for MCMC in general may have significant downsides for ABI. Extremely flat priors, for instance, are often used for MCMC, but may hinder ABI by training the networks on too many unrealistic examples. Conversely, overly narrow or misspecified priors lead to ABI failure since the networks are never trained on relevant data. Another consideration is computational efficiency: density-based MCMC (e.g., Hamiltonian Monte Carlo) require priors whose density is fast to evaluate, whereas ABI benefits from priors that can be efficiently sampled from. This distinction becomes relevant, for example, in our Gaussian mixture model. We adopted an order-restricted prior on component means, a parametrization common in \texttt{Stan}. While efficient for density evaluation, it necessitated rejection sampling for ABI. In our example, well-separated components kept rejection rates low, but with less separation or many more components, rejection sampling would become infeasible, and alternative prior specifications would be required.

One of the main strengths of ABI is inference speed. Across our examples, our implementation of ABI with \texttt{BayesFlow} consistently outperformed MCMC sampling with \texttt{Stan} by several orders of magnitude in inference time. This acceleration is important even when applying a model to a single dataset, since simulation-based validation procedures such as SBC \citep{talts_validating_2018,modrak_simulation-based_2023} typically require fitting hundreds or thousands of datasets, which can become prohibitively expensive with MCMC.

However, the speed benefits apply primarily to the inference phase. \emph{Total} model-fitting time may turn out to be substantially longer \citep{burkner2023some}. Generating training data, training the neural networks, and tuning hyperparameters can demand considerable computational resources which add to the cost of ABI. Implementing ABI pipelines also entails developing and testing the simulator, designing and debugging neural architectures, and experimenting with different network configurations to achieve reliable inference, all of which contribute significant overhead. In our case studies, implementing the new neural factorization, debugging the simulator, and testing multiple network configurations to achieve satisfactory calibration required by far the most time and effort. Moreover, training and inference times depend strongly on network architecture. For example, normalizing flows are typically slower to train than flow-matching architectures \citep{lipman2022flow}, whereas inference with normalizing flows tends to be faster, particularly on CPUs, than flow-matching. These observations underscore that, although ABI can offer substantial gains during inference, it does not necessarily outperform MCMC in overall efficiency in every use case.

Network architectures influence not only speed of training and inference, but also computational faithfulness of the networks. Inappropriate architectures (e.g., using permutation-invariant Deep Sets on time-series data) can distort the model structure, but even within appropriate architectural families, network accuracy is highly sensitive to hyperparameters such as the number of layers, hidden units, activation functions, etc. Simpler networks may underfit and fail to capture relevant patterns in the data; overly complex networks may overfit, reproducing noise from the training data and producing unstable outputs. We observed the latter phenomenon in our first case study, where a deep classification network (12 fully connected layers with ReLU activations) exhibited visibly ``jittery'' output, likely reflecting excessive capacity of the network \citep{novak2018sensitivity}. This setting was the result of our iterative experimentation rather than a principled search for the best hyperparameters. A more principled approach would be using some sort of tuning procedure; for example, Bayesian optimization \citep{snoek2012practical} or population-based training \citep{jaderberg2017population}.

\section*{Acknowledgments}

Paul Bürkner acknowledges support of the DFG Collaborative Research Center 391 (Spatio-Temporal Statistics for the Transition of Energy and Transport) -- 520388526. Paul Bürkner further acknowledges support of the Deutsche Forschungsgemeinschaft (DFG, German Research Foundation) Projects 508399956 and 528702768.

\bibliography{bibliography}
\end{document}

%% file: figures/dag.tex
\node[circle, minimum size = 8mm, thick, draw = black, node distance = 8mm, line width = 1pt] (theta) at (0,0) {$\theta$};
\node[draw, minimum size = 8mm, thick, draw = black, node distance = 8mm, line width = 1pt] (z1) at (-2,-1.5) {$z_1$};
\node[draw, minimum size = 8mm, thick, draw = black, node distance = 8mm, line width = 1pt] (z2) at (0,-1.5) {$z_2$};
\node[draw, minimum size = 8mm, thick, draw = black, node distance = 8mm, line width = 1pt] (z3) at (2,-1.5) {$z_3$};
\node[circle, minimum size = 8mm, thick, draw = black, node distance = 8mm, line width = 1pt, fill = lightgray] (y1) at (-2,-3) {$y_1$};
\node[circle, minimum size = 8mm, thick, draw = black, node distance = 8mm, line width = 1pt, fill = lightgray] (y2) at (0,-3) {$y_2$};
\node[circle, minimum size = 8mm, thick, draw = black, node distance = 8mm, line width = 1pt, fill = lightgray] (y3) at (2,-3) {$y_3$};
\draw[->,thick] (theta) -- (z1);
\draw[->,thick] (theta) -- (z2);
\draw[->,thick] (theta) -- (z3);
\draw[->,thick] (z1) -- (y1);
\draw[->,thick] (z2) -- (y2);
\draw[->,thick] (z3) -- (y3);
